\documentclass[lettersize,journal]{IEEEtran}
\usepackage{amsmath,amsfonts}
\usepackage{algorithmic}
\usepackage{algorithm}
\usepackage{array}
\usepackage[caption=false,font=normalsize,labelfont=sf,textfont=sf]{subfig}
\usepackage{textcomp}
\usepackage{stfloats}
\usepackage{url}
\usepackage{verbatim}
\usepackage{graphicx}
\usepackage{cite}
\usepackage[colorlinks, linkcolor=blue]{hyperref}
\usepackage{multirow}
\usepackage{booktabs}
\usepackage{colortbl}
\usepackage{xcolor}
\usepackage{bm}
\usepackage[table]{xcolor}

\definecolor{aliceblue}{rgb}{0.94, 0.97, 1.0}
\begin{document}

\title{Memory-guided Prototypical Co-occurrence Learning for Mixed Emotion Recognition}

\author{Ming Li, Yong-Jin Liu,~\IEEEmembership{Senior Member,~IEEE}, Fang Liu, Huankun Sheng, Yeying Fan, Yixiang Wei, Minnan Luo, Weizhan Zhang, Wenping Wang,~\IEEEmembership{Fellow,~IEEE}

\IEEEcompsocitemizethanks {\IEEEcompsocthanksitem Ming Li, Yong-Jin Liu, Huankun Sheng, Yeying Fan, and Yixiang Wei are with MOE-Key Laboratory of Pervasive Computing, Department of Computer Science and Technology, Tsinghua University, Beijing, China. Yong-Jin Liu is the corresponding author (e-mail: mingli\_thu, liuyongjin, shenghk, fanyeying@tsinghua.edu.cn, yx-wei22@mails.tsinghua.edu.cn). 
\IEEEcompsocthanksitem Fang~Liu is with the Key State Laboratory of Media Convergence and Communication, Communication University of China, Beijing 100024, China (e-mail: fangliu@cuc.edu.cn).
\IEEEcompsocthanksitem Minnan Luo and Weizhan Zhang are with the School of Computer Science and Technology, and Shaanxi Province Key Laboratory of Big Data Knowledge Engineering, Xi’an Jiaotong University, Xi’an, Shaanxi 710049, China (e-mail: minnluo, zhangwzh@xjtu.edu.cn).
\IEEEcompsocthanksitem Wenping Wang is with the Department of Computer Science and Computer Engineering, Texas A\&M University, College Station, TX 77840 USA (e-mail: wenping@tamu.edu).}
\thanks{This work was supported by the Natural Science Foundation of China (U2336214).}
}
\markboth{Journal of \LaTeX\ Class Files,~Vol.~14, No.~8, August~2021}%
{Shell \MakeLowercase{\textit{et al.}}: A Sample Article Using IEEEtran.cls for IEEE Journals}

\maketitle

\begin{abstract}
Emotion recognition from multi-modal physiological and behavioral signals plays a pivotal role in affective computing, yet most existing models remain constrained to the prediction of singular emotions in controlled laboratory settings. Real-world human emotional experiences, by contrast, are often characterized by the simultaneous presence of multiple affective states, spurring recent interest in mixed emotion recognition as an emotion distribution learning problem. Current approaches, however, often neglect the valence consistency and
structured correlations inherent among coexisting emotions. To address this limitation, we propose a Memory-guided Prototypical Co-occurrence Learning (MPCL) framework that explicitly models emotion co-occurrence patterns. Specifically, we first fuse multi-modal signals via a multi-scale associative memory mechanism. To capture cross-modal semantic relationships, we construct emotion-specific prototype memory banks, yielding rich physiological and behavioral representations, and employ prototype relation distillation to ensure cross-modal alignment in the latent prototype space. Furthermore, inspired by human cognitive memory systems, we introduce a memory retrieval strategy to extract semantic-level co-occurrence associations across emotion categories. Through this bottom-up hierarchical abstraction process, our model learns affectively informative representations for accurate emotion distribution prediction. Comprehensive experiments on two public datasets demonstrate that MPCL consistently outperforms state-of-the-art methods in mixed emotion recognition, both quantitatively and qualitatively.

\end{abstract}


\section{Introduction}

\IEEEPARstart{A}{ffective} Intelligence seeks to equip machines with the ability to perceive and understand human emotions, a crucial step toward enabling natural and empathetic human-computer interaction. Achieving this goal requires both precise emotion recognition and a deeper comprehension of affective states.
Emotions are psychophysiological responses to external stimuli, and their accurate decoding has significant implications for fields such as healthcare and education \cite{wang2023sparse,wu2020electroencephalographic,liu2023brain}. Current research in emotion modeling primarily follows two theoretical frameworks: (1) discrete models, which classify emotions into a set of basic categories \cite{plutchik2003emotions}; and (2) dimensional models, which represent emotions as points within a continuous affective space, such as the Valence-Arousal (VA) model \cite{russell1980circumplex} and the Valence-Arousal-Dominance (VAD) model \cite{mehrabian1996pleasure}. Due to the inherent complexity and nuanced mechanisms underlying emotions, most studies have been conducted in controlled laboratory settings \cite{fu2022multi,li2022neurophysiological}. While these efforts have generated high-quality datasets (e.g., \cite{liu2017real,kosti2019context,zhao2021affective}) and effective recognition models (e.g., \cite{wang2025learning,10976537,cheng2024emotion}), they typically simplify emotions into mutually exclusive, singular states. In real-world scenarios—such as ambiguous situations, social interactions, or conflicting events—humans often experience multiple emotions simultaneously, a phenomenon referred to as {\it mixed emotions}.

Mixed emotions are defined as the co-activation of two or more emotional states \cite{williams2002can}. Growing psychological and neuroscientific evidence indicates that human affective experience is rarely singular; rather, it involves the interplay of multiple emotions, leading to richer and more complex psychological profiles \cite{oh2022specificity,zhao2020multi}. This poses a significant challenge for computational methods, as it requires not only identifying which emotions are present, but also quantifying their respective intensities. Physiologically, mixed emotions involve coordinated activations across distributed brain regions, exhibiting intricate spatiotemporal dynamics \cite{man2017hierarchical,berridge2019affective}. As a result, mixed emotion recognition has emerged as a growing research focus \cite{kreibig2017understanding}, representing a shift from basic emotion classification toward more realistic and nuanced models of human affect. This advancement is essential for uncovering the neural underpinnings and computational principles of complex emotional experiences.

Mixed emotion recognition can be approached through multi-label classification \cite{kang2025beyond}, which identifies the presence of emotions, or through emotion distribution learning (EDL), which estimates the intensity of each emotion. EDL has been used to model emotional diversity in images \cite{xu2025multiple} and to capture subtle emotional semantics in text \cite{zhang2018text}. However, single-modal data often fail to fully capture the complexity of mixed emotions. Recent studies have therefore turned to multimodal approaches that combine physiological and behavioral signals, supported by publicly available datasets. For example, EDLConV Network \cite{shu2022emotion} uses peripheral physiological signals—including electrocardiogram (ECG), heart rate (HR), galvanic skin response (GSR), and skin temperature (SKT)—to decode the distribution of basic emotions. EmotionDict \cite{liu2023emotion} further integrates electroencephalogram (EEG), peripheral signals, and facial videos to decompose mixed emotions into basic emotional components in a latent space. While these methods capture the presence and intensity of individual emotions, they largely treat emotion categories as independent, overlooking the intrinsic relationships among them. Helo \cite{zheng2025helo} addresses this issue by modeling label correlations and exploiting cross-modal heterogeneity. However, it still relies on statistical dependencies derived from label distributions, without incorporating valence consistency and mutual exclusivity priors inherent in the affective space.

Psychological studies \cite{moore2022taking} indicate that emotions exhibit structured relationships based on valence. Emotions sharing the same valence (e.g., both positive or both negative) are often positively correlated and tend to manifest simultaneously, a phenomenon known as {\it co-occurrence} \cite{VANSTEELANDT2005325}. In contrast, emotions with opposite valences typically engage in competitive or inhibitory interactions, making their simultaneous activation less likely \cite{primoceri2023cross}. This reflects the fact that same-valence emotions are proximal in affective space, whereas opposite-valence emotions are more mutually exclusive \cite{malezieux2023neural}. Mixed emotions thus arise from combinations of basic emotions governed by these co-occurrence patterns. Nevertheless, most existing computational approaches overlook this intrinsic structure \cite{yang2021circular,zhao2019text} and instead focus on holistic and salient feature representations for distribution prediction. As a result, they fail to capture the internal co-occurrence relationships present in raw multimodal data, hindering the modeling of emotion interaction relationships and latent compositional structures. Moreover, each basic emotion within a mixed state is associated with specific feature activations \cite{peng2024carat}, yet raw data often consist of low-level statistical features entangled with noise, rather than semantically disentangled representations. In such a feature-entangled space, directly modeling co-occurrence patterns may capture superficial data correlations rather than psychologically meaningful affective structures.

Inspired by the psychological principle of emotion co-occurrence, we propose a novel {\it Memory-guided Prototypical Co-occurrence Learning} (MPCL) framework to model the semantic structural associations among emotions. Our framework is organized into three interconnected stages. (1) {\it Multimodal Feature Extraction and Fusion}: this stage employs a multi-scale associative memory strategy to fuse multimodal physiological and behavioral signals, capturing intrinsic correlations and complementary information. (2) {\it Prototypical Alignment and Co‑occurrence Learning}: we extract fine-grained emotion prototypes to construct an emotion memory bank, yielding semantically rich embeddings by reconstructing physiological and behavioral features as weighted combinations of these prototypes. Prototype Relation Distillation (PRD) is employed to enforce structural alignment within the prototype-relational space, thereby ensuring semantic consistency across heterogeneous modalities. In this stage, drawing on human association memory mechanisms \cite{bonner2021object,potter2012conceptual}, we further present Prototypical Co-occurrence Learning (PCL). PCL employs a memory retrieval strategy to aggregate similar samples within semantic neighborhoods, realizing semantic-level prototype co-occurrence, complemented by a contrastive learning objective to ensure cross-modal consistency. (3) {\it Hierarchical Semantic Compression and Distribution Prediction}: we derive highly condensed emotion representations via a bottom-up hierarchical abstraction strategy, enabling precise mixed emotion distribution prediction. 

The main contributions of this paper include:

\begin{itemize}
    \item We propose a multi-scale associative memory fusion strategy to capture intrinsic emotion correlations in multimodal physiological data, and introduce a content-addressable memory mechanism to construct fine-grained prototype memory banks, generating semantically rich and structurally consistent representations.

    \item Inspired by human cognitive memory mechanisms, we design a memory retrieval strategy to extract prototype-level semantic co-occurrences, effectively uncovering latent structural associations among distinct emotions.

    \item We develop a hierarchical semantic compression method that progressively abstracts affective representations through bottom-up processing, enabling accurate emotion distribution learning. Extensive experiments on two public benchmarks demonstrate that our approach achieves state-of-the-art performance in mixed emotion recognition, both quantitatively and qualitatively.
    
\end{itemize}

\section{Related work}
\subsection{Multimodal Emotion Distribution Learning}

Emotion recognition plays a central role in affective intelligence and commonly processes inputs from multiple modalities, including visual, auditory, textual, and physiological signals \cite{11150759}. Among these, physiological signals are less susceptible to intentional control or masking, offering a more direct reflection of genuine emotional states. In particular, electroencephalogram (EEG) has been widely adopted in affective computing due to its millisecond temporal resolution in capturing cortical dynamics \cite{9979692}. Compared to unimodal methods, multimodal emotion recognition (MER) achieves superior performance by integrating complementary information across modalities, leading to the development of diverse fusion strategies \cite{li2025tracing,li2023decoupled}.

MER can be broadly categorized into single-label and multi-label recognition. While the majority of existing studies focus on single-label recognition—assuming each data sample is associated with one dominant emotion \cite{10731546}—multi-label recognition aims to detect multiple coexisting emotions. The latter aligns more closely with psychological reality, where emotional experiences are often ambiguous and complex; individuals in real-world situations frequently undergo mixed emotions, characterized by the simultaneous presence of multiple affective states \cite{grossmann2017mixed}. For instance, MAGDRA \cite{li2024magdra} models cross-modal interactions among visual, audio, and textual data, and incorporates a reinforced multi-label detection module to learn label correlations. Similarly, SeMuLPCD \cite{anand2023multi} employs a multimodal peer collaborative distillation mechanism to transfer complementary knowledge from unimodal networks to a fusion network for multi-label emotion prediction.


Traditional multi-label emotion recognition typically formulates the task as multiple independent binary classification problems, thereby only identifying the presence or absence of each emotion. In contrast, multimodal emotion distribution learning (EDL) extends this by further quantifying the intensity of each emotional state. This perspective resonates with the General Psycho-evolutionary Theory \cite{plutchik1980general}, which posits that a set of basic emotion prototypes exists with varying intensities, and that complex emotions arise from mixtures of these prototypes. Following this notion, Shu et al. \cite{shu2022emotion} designed the EDLConV model, which leverages feature correlations and temporal dynamics from peripheral physiological signals to decode mixed emotion distributions. EmotionDict \cite{liu2023emotion} integrates peripheral physiology, EEG, and facial video features, using a dictionary learning module to decompose them into weighted combinations of basic emotional components. MEDL \cite{jia2022multimodal} learns emotion distributions and label correlations separately for video and audio modalities, and enforces consistency between cross-modal correlation matrices to predict the final emotion distribution. More recently, HeLo \cite{zheng2025helo} incorporates Optimal Transport theory to facilitate heterogeneous interactions between physiological and behavioral signals, and adopts a label correlation-driven attention mechanism for distribution prediction.

Although these methods effectively exploit cross-modal complementarity and label correlations, they primarily rely on data-driven statistical associations to model inter-label dependencies. However, psychological studies \cite{chou2022exploiting} confirm that emotions sharing the same valence are more likely to co-activate than those with opposite valences—a phenomenon known as emotion co-occurrence. Therefore, in this paper, we aim to incorporate this psychological prior to guide the learning of structured correlations among different emotional states.

\subsection{Prototype-based Representation Learning}

Prototype learning aims to identify a representative center for each class, referred to as a prototype or centroid, and has been widely studied in pattern recognition and machine learning \cite{liu2001evaluation}. Early prototype-based methods predominantly relied on geometric distance metrics. For example, k-nearest neighbors (KNN) \cite{dudani1976distance} determines the class of a query sample by its distance to training instances, while k-means clustering \cite{caron2018deep} and learning vector quantization (LVQ) \cite{kohonen2002self} assign samples to the nearest cluster centroids. Although such approaches reduce dependency on the full training set during inference, distance-based measures in high-dimensional feature spaces are often vulnerable to the curse of dimensionality and may fail to capture semantically meaningful relations.

With the advancement of deep learning, prototype learning has emerged as a powerful framework for automatically learning discriminative representations. Initially proposed for few-shot image classification, Snell et al. \cite{snell2017prototypical} introduced prototypes as class-wise representatives in a learned embedding space, where classification is performed by measuring the distance between a query sample and each prototype. This idea was later extended to few-shot semantic segmentation \cite{dong2018few}. Inspired by these visual applications, prototype learning has been successfully adapted to emotion recognition. For instance, MapleNet \cite{11157707} employs a shared encoder to align textual and visual features with emotion-specific prototypes, achieving cross-modal semantic alignment. PR-PL \cite{zhou2023pr} uses prototype representations to model the latent semantic structure of emotion categories in EEG signals, obtaining subject-invariant and generalizable affective representations.

Another important line of research integrates prototypes into contrastive learning frameworks. While conventional instance-wise contrastive methods focus on discriminating between augmented views of the same sample, they often overlook higher-level semantic structures. To address this, prototypical contrastive learning \cite{li2020prototypical} introduces cluster centroids as prototypes and encourages samples to align with their corresponding prototypes, thereby injecting semantic structures into the learned representation space. Likewise, ProtoCLIP \cite{10339644} uses prototypes as stable anchors to cluster semantically similar instances, improving grouping performance in vision–language contrastive learning. In the emotion recognition domain, SRMCL \cite{bao2024boosting} proposes supervised prototype memory contrastive learning (PMCL) to enhance intra-class compactness for micro-expression recognition by contrasting current and historical prototypes. SCCL \cite{yang2023cluster} leverages sentiment lexicon–derived coordinates as predefined prototypes to perform cluster-level contrastive learning, capturing fine-grained emotional semantics. EmotionDict \cite{liu2023emotion} incorporates a dictionary learning module that decomposes mixed emotions into linear combinations of basic emotion elements, strengthening the expressiveness of multimodal features.

Despite their effectiveness, existing prototype-based methods in emotion recognition suffer from several limitations. Many rely on predefined prototypes, simple averaging over samples, or implicit basis representations, which are sensitive to noise and prone to prototype drift. This often leads to semantically blurred representations that fail to capture fine-grained affective nuances and intra-class diversity. Furthermore, current cross-modal prototype alignment strategies typically neglect the global relational structure among prototypes, resulting in poor geometric consistency across modalities in the shared semantic space. In contrast, our approach presented in this paper reconstructs multimodal features as weighted combinations of fine-grained emotion prototypes to obtain semantically rich embeddings, and enforces cross-modal structural consistency through prototype relation distillation, thereby preserving both local discriminability and global relational semantics.

\subsection{Associative Memory and Modern Hopfield Networks}

Associative memory networks are computational models that emulate the human brain’s ability to store and retrieve patterns, allowing for the recovery of complete memories from partial or noisy inputs \cite{mceliece2003capacity}. The classic Hopfield network, introduced in the 1980s, is an energy-based fully connected neural network that stores and retrieves information through a symmetric weight matrix \cite{hopfield1984neurons}. Recognized for its stability and associative recall properties, it has found broad application in pattern recognition tasks \cite{zhang2022out}. However, the storage capacity of traditional binary Hopfield networks is notably limited. To address this, researchers have proposed modern Hopfield networks (MHNs), also referred to as dense associative memories (DAMs), primarily by redesigning the energy function. Krotov and Hopfield \cite{krotov2016dense} generalized the neuron interaction from $F(x) = x^2$ to $F(x) = x^n$, achieving a storage capacity that scales polynomially with dimension as $d^{n-1}$. Demircigil et al. \cite{demircigil2017model} further improved capacity to an exponential scale using $F(x) = \exp(x)$. More recently, Ramsauer et al. \cite{ramsauerhopfield} extended the exponential energy formulation to continuous states, introducing the continuous modern Hopfield network. Its retrieval update rule permits single-step memory recall and is mathematically equivalent to the attention mechanism in Transformers, enabling seamless integration into deep learning frameworks for tasks such as feature aggregation, memory-augmented learning, prototype representation, and structured attention.

Continuous MHNs have demonstrated strong performance across diverse applications, including immune repertoire classification \cite{widrich2020modern}, chemical reaction prediction \cite{seidl2022improving}, and reinforcement learning \cite{widrich2021modern}, owing to several key properties. First, their associative memory mechanism supports effective multimodal fusion by modeling interactions between query patterns (states) and stored memory patterns (keys), thereby capturing intrinsic cross-modal correlations and complementary information. Second, MHNs possess notable feature aggregation capabilities, summarizing input representations within a memory space via static, learnable state patterns. Moreover, MHNs facilitate memory representation learning by parameterizing the memory space as a trainable matrix — often constructed from training samples, reference sets, or prototypes — that can be optimized to encode the underlying structure of the data.

Recently, MHNs have also been incorporated into contrastive learning frameworks to address the explaining-away problem. Methods such as CLOOB \cite{furst2022cloob} and CLOOME \cite{sanchez2023cloome} employ MHNs to enrich the covariance structure of the original embeddings. By substituting original features with their retrieved counterparts from the memory, these approaches reinforce frequently co-occurring feature patterns. However, existing retrieval strategies are largely confined to feature-level similarity aggregation, often overlooking deeper, semantically structured relationships. In contrast, our work leverages the memory retrieval mechanism of MHNs to model cross-modal co-occurrence consistency within a memory-augmented semantic space, thereby capturing richer and more structured affective associations.

\section{Methods}

\begin{figure*}[!t]
\centering
\includegraphics[width=\linewidth]{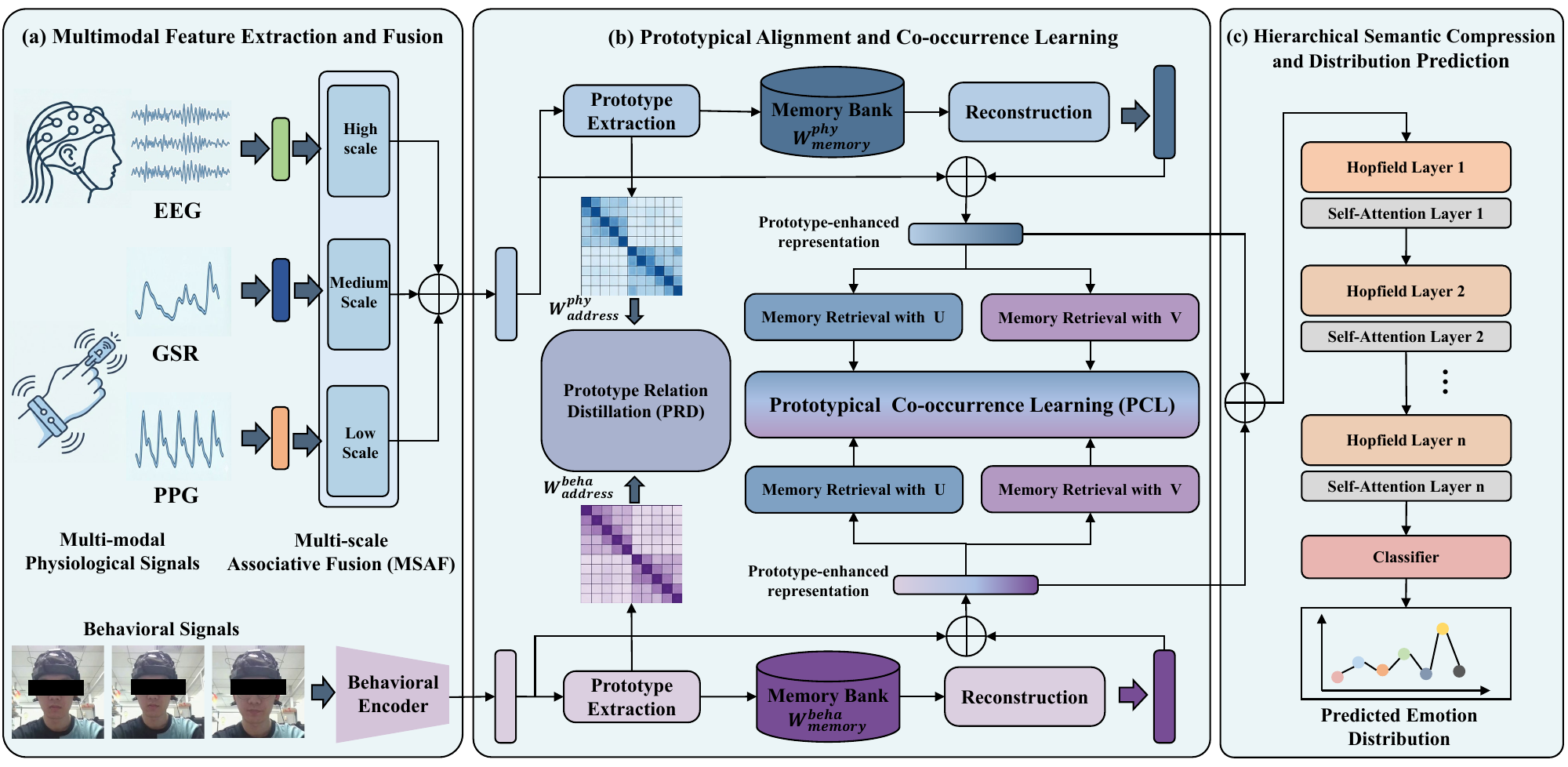}
\caption{Overview of the MPCL Framework, which consists of three stages. (a) Multimodal Feature Extraction and Fusion: the Multi‑Scale Associative Fusion (MSAF) module fuses multi-modal physiological signals, while a separate behavioral encoder extracts structured behavioral feature. (b) Prototypical Alignment and Co‑occurrence Learning: emotion prototype memory banks are constructed from both modalities. The Prototype Relation Distillation (PRD) module enforces cross‑modal structural alignment, and the Prototypical Co‑occurrence Learning (PCL) module captures semantic‑level co‑occurrence patterns through memory retrieval. (c) Hierarchical Semantic Compression and Distribution Prediction: the Hierarchical Semantic Compression (HSC) module abstracts affective representations via a bottom‑up strategy, followed by a classifier that outputs the final emotion distribution.}
\label{fig_1}
\end{figure*}

\subsection{Preliminaries: Energy-Based Hopfield Networks}

\subsubsection{Energy Function}
The energy-based continuous modern Hopfield network \cite{ramsauerhopfield} serves as the unified theoretical foundation for our proposed MPCL framework. Unlike the traditional attention mechanism in Transformers, MHNs allow the iterative optimization process to be interpreted as energy minimization. Formally, consider a state pattern (query) $\bm{\xi} \in \mathbb{R}^d$ and a set of $M$ stored patterns (memory) $\mathbf{X} = (\bm{x}_1, \dots, \bm{x}_M) \in \mathbb{R}^{d \times M}$. The network dynamics are governed by an energy function $E$, defined via the Log-Sum-Exp (lse) function as:

\begin{equation}
\begin{split}
E(\bm{\xi}) &= -\text{lse}(\beta, \mathbf{X}^T \bm{\xi}) + \frac{1}{2} \bm{\xi}^T \bm{\xi} + C_0 \\
&= -\frac{1}{\beta} \log \left( \sum_{j=1}^M \exp(\beta \bm{x}_j^T \bm{\xi}) \right) + \frac{1}{2} \bm{\xi}^T \bm{\xi} + C_0,
\end{split}
\label{eq:1}
\end{equation}
where $\beta > 0$ is the inverse temperature parameter that controls the sharpness of the energy landscape, and $C_0$ is a normalization constant. The first term promotes alignment of the state vector $\bm{\xi}$ with the stored patterns (associative recall), while the quadratic term acts as a regularizer that constrains the norm of the state vector.

\subsubsection{Update Rule and Unified Hopfield Operator}
The update rule that minimizes the energy function in Eq. (\ref{eq:1}) in a single step is mathematically equivalent to the well-known attention mechanism in Transformers \cite{vaswani2017attention, ramsauerhopfield}:
\begin{equation}
\bm{\xi}^{\text{new}} = \mathbf{X} \, \text{softmax}(\beta \mathbf{X}^T \bm{\xi}).
\label{eq:2}
\end{equation}

To provide a modular and concise description of our MPCL framework, we generalize this update rule into a matrix formulation and abstract it into a unified Hopfield operator. Given state patterns \( \mathbf{R} \in \mathbb{R}^{S \times d_k} \) as queries, stored patterns \( \mathbf{Y} \in \mathbb{R}^{N \times d_k} \) as keys, and pattern projections \( \mathbf{V} \in \mathbb{R}^{N \times d_v} \) as values, the operator is defined as:
\begin{equation}
\text{Hopfield}(\mathbf{R}, \mathbf{Y}, \mathbf{V}, \beta) := \text{softmax}\left( \beta \mathbf{R} \mathbf{Y}^T \right) \mathbf{V}.
\label{eq:3}
\end{equation} 

\subsubsection{Physical Interpretation} From a dynamical systems viewpoint, this operator implements content-addressable memory retrieval, in which patterns stored in $\mathbf{Y}$ act as attractors (or metastable states) within the energy landscape. Although mathematically equivalent to Transformer attention, this update rule is rigorously proven to converge to stationary points of the energy function, i.e., local minima or saddle points~\cite{ramsauerhopfield}. Within our framework, by configuring $\mathbf{R}$, $\mathbf{Y}$, $\mathbf{V}$ and $\beta$, we leverage this operator to realize four key functional modules: cross-modal association (Section \ref{sec:multi_scale_associative_fusion}), prototype memory bank construction (Section \ref{sec:EmotionMemoryBank}), prototype co-occurrence (Section \ref{sec:Prototypical_Cooccurrence_Learning}), and semantic compression (Section \ref{sec:Hierarchical_Semantic_Compression}).

\subsection{Task Formulation and Framework Overview}

\subsubsection{Task Formulation}
We formulate multimodal mixed emotion recognition as an emotion distribution learning problem. Let $\mathcal{D} = \{(\bm{x}_i, \bm{y}_i)\}_{i=1}^T$ denote a dataset containing $T$ samples, where each sample $\bm{x}_i = \{\bm{x}_i^{\text{phy}}, \bm{x}_i^{\text{beha}}\}$ consists of physiological signals and behavioral observations. Specifically, $\bm{x}_i^{\text{phy}} \in \mathbb{R}^{C_{\text{phy}} \times d_{\text{phy}}}$ represents the set of physiological signals (e.g., EEG, GSR, or ECG), and $\bm{x}_i^{\text{beha}} \in \mathbb{R}^{C_{\text{beha}} \times d_{\text{beha}}}$ corresponds to behavioral data such as facial videos. Here, $C_{(\cdot)}$ and $d_{(\cdot)}$ denote the number of channels and feature dimension per modality, respectively. The label $\bm{y}_i \in \mathbb{R}^E$ is a probability distribution over $E$ emotion categories, satisfying the normalization constraint $\sum_{j=1}^E y_{i,j} = 1$. Our objective is to learn a mapping function $\mathcal{F}(\bm{x}; \bm{\theta})$ parameterized by $\bm{\theta}$ that minimizes the divergence between the predicted distribution $\hat{\bm{y}}_i$ and the ground truth $\bm{y}_i$.

\subsubsection{Framework Overview}

As illustrated in Fig.~\ref{fig_1}, our proposed MPCL framework consists of three core stages: 1) {\bf Multimodal Feature Extraction and Fusion}: The {\it Multi-Scale Associative Fusion} (MSAF) module encodes and integrates multimodal physiological signals into a unified embedding through an associative memory mechanism. Concurrently, a behavior encoder maps input raw signals into structured behavioral embeddings. 2) {\bf Prototypical Alignment and Co‑occurrence Learning}: Emotion prototype memory banks are constructed from both physiological and behavioral embeddings. {\it Prototype Relation Distillation} (PRD) is utilized to perform prototype-based denoising and to enforce cross-modal structural alignment between the two modalities. Subsequently, the {\it Prototypical Co-occurrence Learning} (PCL) module reinforces cross-modal semantic co-occurrence consistency via memory retrieval and contrastive learning. 3) {\bf Hierarchical semantic compression and distribution prediction}: The {\it Hierarchical Semantic Compression} (HSC) module abstracts high-level affective  concepts from the co-occurrence-enhanced representations. Finally, a classifier predicts the emotion distribution for each sample.

\subsection{Multimodal Feature Extraction and Fusion}
\label{sec:multi_scale_associative_fusion}

\begin{figure}[!t]
\centering
\includegraphics[width=\linewidth]{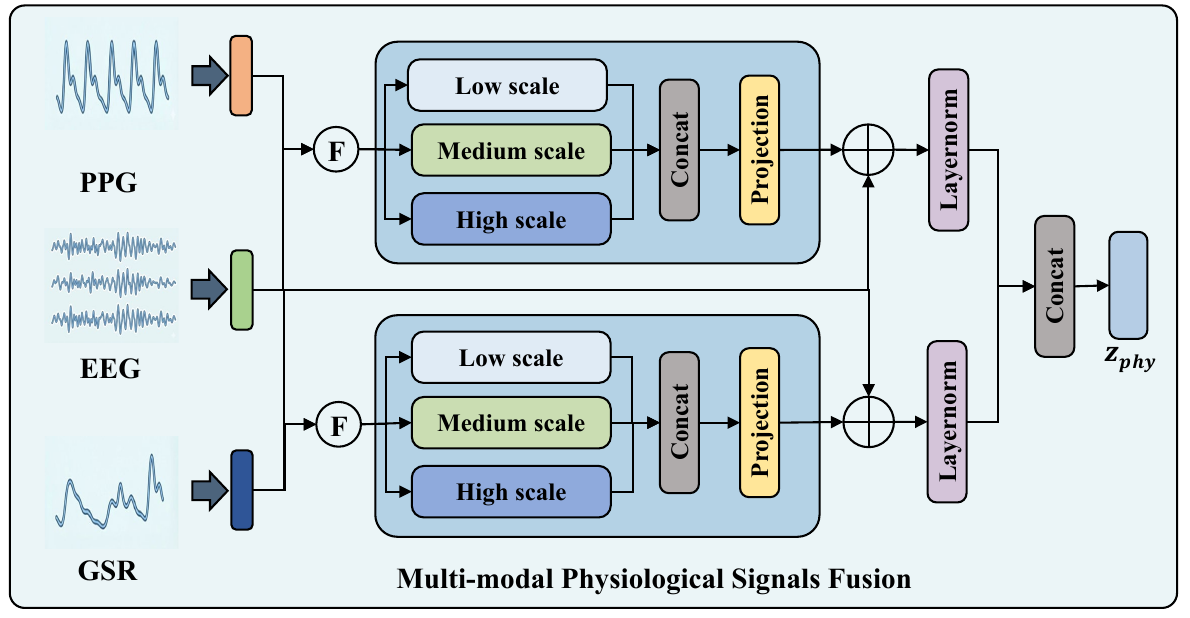}
\caption{Fusion of multimodal physiological signals. EEG serves as the primary modality to integrate complementary information from auxiliary modalities (PPG and GSR) via a multi-scale associative memory mechanism. Three distinct scaling factors $\mathcal{B} = \{\beta_{\text{low}}, \beta_{\text{mid}}, \beta_{\text{high}}\}$ are employed to modulate the granularity of information aggregation.}
\label{fusion}
\end{figure}

Our framework processes both physiological and behavioral signals. 
Distinct from externally observable behavioral signals (such as facial videos or accelerometer data), physiological signals capture a broad range of internal biological activities—from central nervous activity (e.g., EEG) to cardiovascular (e.g., ECG or photoplethysmogram, PPG), electrodermal (e.g., GSR or electrodermal activity, EDA) and muscular (e.g., electromyography, EMG) dynamics. These continuous time series signals collectively reflect the subject’s intrinsic physiological state. To exploit their inherent synergy, we prioritize the fusion of these modalities, as illustrated in Fig.~\ref{fusion}.

After standard pre-processing (Section \ref{subsec:data-proc}), we denote the physiological signals as $\bm{x}^{\text{phy}} \in \mathbb{R}^{C_{\text{phy}} \times d_{\text{phy}}}$ and the behavioral signal as $\bm{x}^{\text{beha}} \in \mathbb{R}^{C_{\text{beha}} \times d_{\text{beha}}}$. Based on the acquisition characteristics and typical precision of the recording devices, we further partition the physiological set $\bm{x}^{\text{phy}}$ into a primary modality $\bm{x}^{\text{pri}} \in \mathbb{R}^{C_{\text{pri}} \times d_{\text{pri}}}$ (e.g., EEG) and a set of auxiliary modalities $\{\bm{x}^{m}\}_{m \in \mathcal{M}_{\text{aux}}}$, where $\mathcal{M}_{\text{aux}}$ indexes the auxiliary signals (e.g., $\mathcal{M}_{\text{aux}} = \{\text{GSR}, \text{PPG}\}$). 

To mitigate feature discrepancies across heterogeneous modalities, we employ modality-specific encoders $f_{(\cdot)}$ to project all raw inputs into a unified latent space with feature dimension $D$ and a standardized sequence length $C$. The projection process is formulated as:
\begin{equation}
\bm{h}^{\text{pri}} = f_{\text{pri}}(\bm{x}^{\text{pri}}),\;
\bm{h}^{m} = f_{m}(\bm{x}^{m}),\;
\bm{h}^{\text{beha}} = f_{\text{beha}}(\bm{x}^{\text{beha}})
\end{equation}
where $\bm{h}^{\text{pri}}, \bm{h}^{m} \in \mathbb{R}^{C \times D}$ are the projected physiological embeddings. For structural symmetry in later cross-modal interaction, the behavioral input is encoded and projected into $\bm{h}^{\text{beha}} \in \mathbb{R}^{K \times D}$, where $K = |\mathcal{M}_{\text{aux}}| \cdot C$, matching the aggregated sequence length of the physiological modality.

To fully exploit the complementary information across modalities, we propose a multi-scale fusion strategy governed by a set of scaling parameters $\mathcal{B} = \{\beta_{\text{low}}, \beta_{\text{mid}}, \beta_{\text{high}}\}$ (Fig.~\ref{fusion}). Unlike standard attention that uses a fixed scaling factor mainly for gradient stability, our strategy leverages $\beta$ to modulate the granularity of information aggregation, framing multimodal fusion as an energy minimization process. A smaller $\beta$ promotes global context aggregation while suppressing modality-specific noise, whereas a larger $\beta$ enforces precise local‑feature alignment. For each scale $\beta \in \mathcal{B}$, the Hopfield operator is applied as:
\begin{equation}
    \bm{h}_{\beta}  ^m = \text{Hopfield}(\bm{h}^{\text{pri}}, \bm{h}^m, \bm{h}^m, \beta)
\end{equation}
Granularity-specific representations are then concatenated along the feature dimension and projected back to the latent space $D$. Defining the concatenated feature as $\bm{h}_{\text{cat}}^m \in \mathbb{R}^{C \times (|\mathcal{B}| \cdot D)}$, the final fused representation $\bm{z}_{\text{phy}}^m \in \mathbb{R}^{C \times D}$ is obtained via a learnable projection $\mathbf{W}_{\text{proj}} \in \mathbb{R}^{(|\mathcal{B}| \cdot D) \times D}$ and a residual connection:
\begin{equation}
    \bm{h}_{\text{cat}}^m = \text{Concat}\left( \{ \bm{h}_{\beta}^m \mid \beta \in \mathcal{B} \} \right)
\end{equation}
\begin{equation}
    \bm{z}^{\text{phy}, m} = \bm{h}^{\text{pri}} + \bm{h}_{\text{cat}}^m \mathbf{W}_{\text{proj}}
\end{equation}
Finally, we aggregate the fused physiological representations from all auxiliary modalities. The resulting physiological and behavioral representations are:
\begin{equation}
    \bm{z}^{\text{phy}} = \text{Concat}\left( \{ \bm{z}^{\text{phy}, m} \mid m \in \mathcal{M}_{\text{aux}} \} \right), \  \bm{z}^{\text{beha}} = \bm{h}^{\text{beha}}
\end{equation}

\subsection{Prototypical Alignment and Co‑occurrence Learning}
\label{sec:Prototypical_Alignment_and_Cooccurrence_Learning}

\begin{figure}[!t]
\centering
\includegraphics[width=\linewidth]{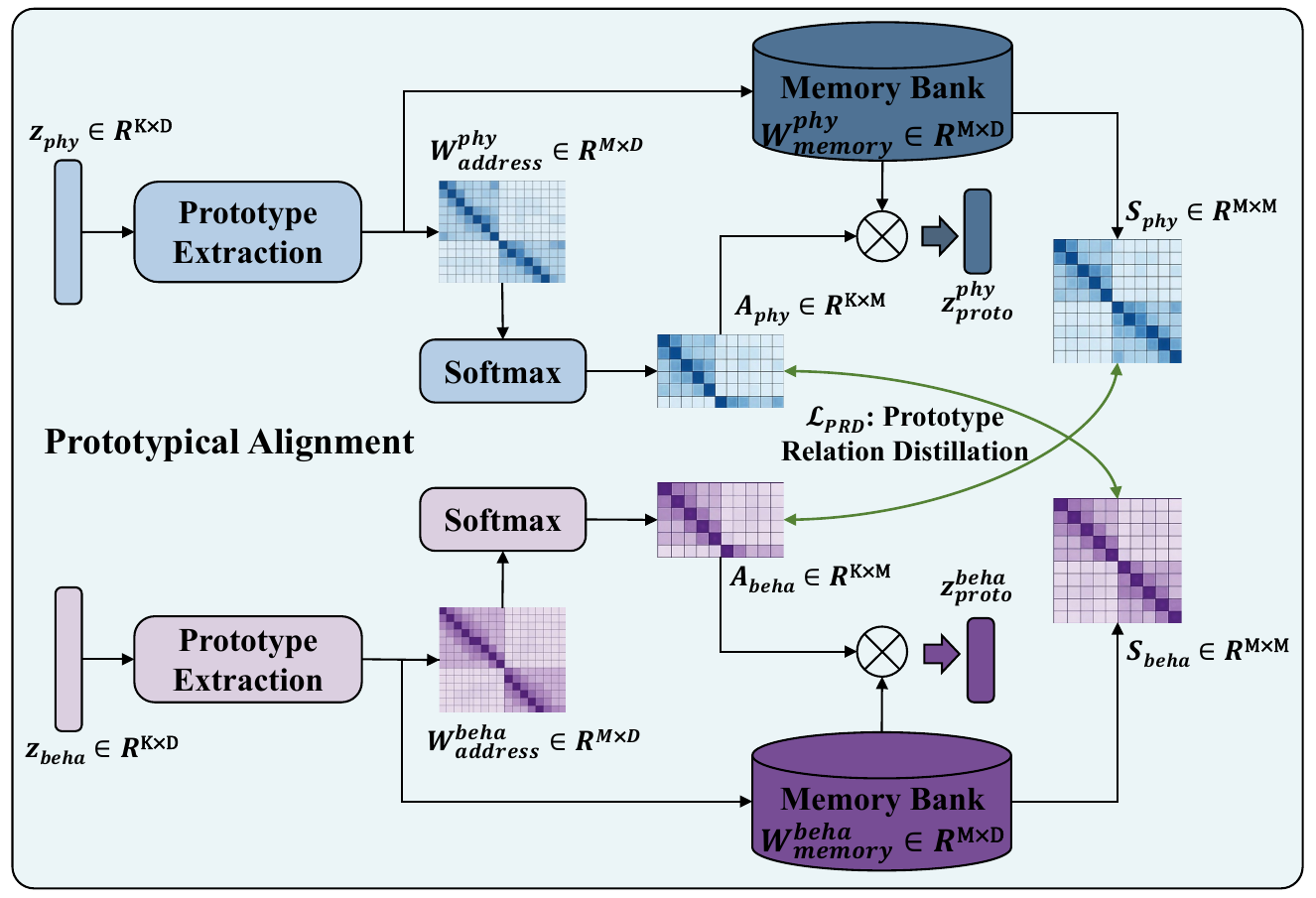}
\caption{Prototype alignment process. Prototype memory banks are first constructed separately for the physiological and behavioral modalities, respectively. A semantics-enriched representation is then obtained as a weighted combination of prototypes from each bank. Meanwhile, the Prototype Relation Distillation (PRD) strategy enforces semantic structural consistency across the two modalities.}
\label{fig_PA}
\end{figure}

While we have obtained fused physiological and behavioral embeddings, these representations essentially remain in a low-level feature space that lacks explicit semantic disentanglement. Directly modeling co-occurrence in such a space risks capturing spurious, data-level correlations rather than psychologically grounded emotional structures. To address this, we introduce a prototypical modern Hopfield network that learns a set of fine-grained emotion prototypes and constructs a global emotion memory bank. As shown in Fig.~\ref{fig_PA}, our goal is to map heterogeneous multimodal signals into a unified, semantics-enhanced prototype space.

\subsubsection{Prototype Memory Bank}
\label{sec:EmotionMemoryBank}

Let $\bm{z} \in \mathbb{R}^{K \times D}$ denote a generic input embedding (either $\bm{z}_{\text{phy}}$ or $\bm{z}_{\text{beha}}$). To learn and extract prototypes, we employ two trainable parameter matrices: one for prototype addressing and the other for prototype storage.
\begin{itemize}
    \item Prototype Address Matrix $\mathbf{W}_{\text{address}} \in \mathbb{R}^{M \times D}$ stores representative signatures used for associative addressing. It defines the latent addressing space of the memory bank, where each row act as a reference pattern that interacts with the input query $\bm{z}$ to compute the prototype assignment weights.
    \item Prototype Memory Matrix $\mathbf{W}_{\text{memory}} \in \mathbb{R}^{M \times D}$ stores the semantic content of each prototype. It provides the semantic basis for feature reconstruction, with each row corresponding to a learned prototype. The input embedding $\bm{z}$ is represented as a weighted aggregation of these prototype contents.
\end{itemize}
Here, $M$ denotes the memory capacity (i.e., the number of prototypes). First, we compute the addressing distribution $\mathbf{A} \in \mathbb{R}^{K \times M}$ as:
\begin{equation}
\mathbf{A} = \text{softmax}(\beta \bm{z} \mathbf{W}_{\text{address}}^{T})
\label{eq:9}
\end{equation}
where $\beta = 1/\sqrt{D}$ is a scaling factor. This distribution represents the weight (or probability) of each prototype, quantifying its contribution to the input sample. Next, we reconstruct the input $\bm{z}$ as a weighted combination of the prototype memories stored in $\mathbf{W}_{\text{memory}}$, yielding the semantics-enriched embedding $\bm{z}_{\text{proto}} \in \mathbb{R}^{K \times D}$:
\begin{equation}
    \bm{z}_{\text{proto}} = \mathbf{A} \mathbf{W}_{\text{memory}} = \text{Hopfield}(\bm{z}, \mathbf{W}_{\text{address}}, \mathbf{W}_{\text{memory}}, \beta)
\end{equation}
Notably, this learned prototype bank is subsequently used to initialize the hierarchical semantic compression module.

\subsubsection{Prototype Relation Distillation (PRD)}

To mitigate cross-modal heterogeneity and enforce semantic structural consistency, we propose a PRD strategy that leverages the extracted prototypes for cross‑modal supervision. First, the addressing weight matrix $\mathbf{A}$ from Eq. (\ref{eq:9}) serves as the student distribution, representing the instance-wise assignment probabilities over all prototypes. Concurrently, we construct a "ground truth" relational structure among the learned prototypes to provide a teacher distribution. The semantic correlation matrix $\mathbf{S} \in \mathbb{R}^{M \times M}$ is defined as:
\begin{equation}
\mathbf{S} = \text{softmax}\left( \frac{\mathbf{W}_{\text{memory}} \mathbf{W}_{\text{address}}^T}{\tau_{\text{dist}}} \right) 
\end{equation}
where $\tau_{\text{dist}}$ is a temperature parameter. Each row $\mathbf{S}_i$ captures the semantic distribution of the $i$-th prototype relative to the entire memory bank, offering structure-aware soft labels for cross-modal distillation.

We then use the topological structure of the dominant prototype from one modality to supervise the addressing distribution of the other modality. For each sample in a mini-batch of size $N$, we first identify the pseudo-label prototype indices based on the maximum activation of the respective teacher modalities:
\begin{equation}
    p^*_i = \mathop{\arg\max}_{j} (\mathbf{A}^{\text{beha}}_{i,j}), \quad
    q^*_i = \mathop{\arg\max}_{j} (\mathbf{A}^{\text{phy}}_{i,j})
\end{equation}
Subsequently, Kullback-Leibler (KL) divergence \cite{kullback1951information} is employed to enforce cross-modal semantic consistency. By minimizing the divergence between the student's addressing distribution and the retrieved teacher topology, the PRD loss is formulated as:
\begin{equation}
    \mathcal{L}_{\text{PRD}} = \frac{1}{2} \sum_{i=1}^{N} \left( D_{\text{KL}}\left( \mathbf{S}^{\text{beha}}_{p^*_i} \parallel \mathbf{A}^{\text{phy}}_i \right) + D_{\text{KL}}\left( \mathbf{S}^{\text{phy}}_{q^*_i} \parallel \mathbf{A}^{\text{beha}}_i \right) \right)
\end{equation}

\subsubsection{Prototypical Co-occurrence Learning (PCL)}
\label{sec:Prototypical_Cooccurrence_Learning}
\begin{figure}[!t]
\centering
\includegraphics[width=\linewidth]{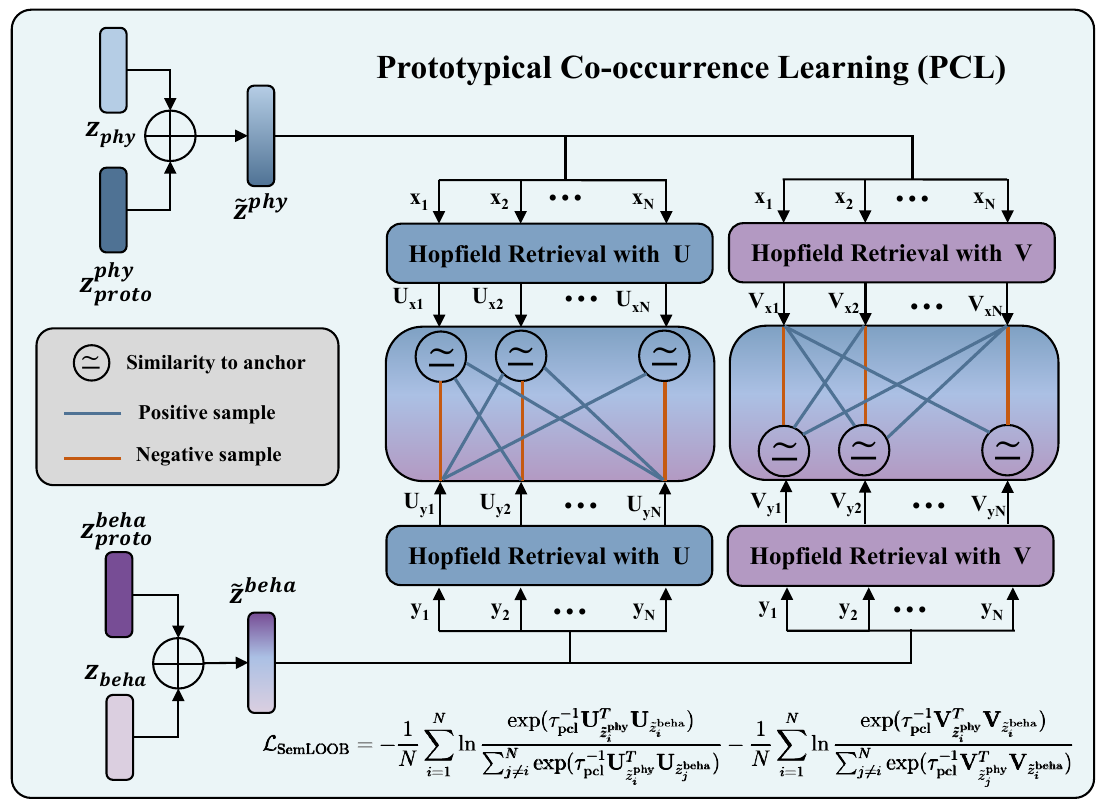}
\caption{Architecture of Prototypical Co‑occurrence Learning (PCL). The physiological embedding $x_i$ and the behavioral embedding $y_i$ retrieve associated representations $U_{x_i}$ and $U_{y_i}$, respectively, via the Hopfield network. Within the physiological embedding space, $U_{x_i}$ (retrieved via the physiological query) serves as an anchor that is contrasted with the positive sample $U_{y_i}$ (retrieved from the corresponding behavioral query) and with negative samples $U_{y_j}$ (retrieved from mismatched behavioral queries, where $j \neq i$). The same procedure is applied symmetrically in the behavioral embedding space $V$.}
\label{fig_PCL}
\end{figure}

To uncover the relational structure of basic emotions within mixed emotional states, we leverage the memory retrieval mechanism of Hopfield networks to amplify semantic co-occurrence patterns. As shown in Fig. \ref{fig_PCL}, this design is inspired by the brain’s ability to extract co-occurrence regularities through visual perception and memory. Specifically, we first augment the raw input features $\bm{z}$ with the learned prototype features $\bm{z}_{\text{proto}}$, yielding prototype-enhanced representations:
\begin{equation}
\tilde{\bm{z}}^{\text{phy}} = \bm{z}^{\text{phy}} + \bm{z}_{\text{proto}}^{\text{phy}}, \quad
\tilde{\bm{z}}^{\text{beha}} = \bm{z}^{\text{beha}} + \bm{z}_{\text{proto}}^{\text{beha}}
\end{equation}
where $\tilde{\bm{z}}^{(\cdot)} \in \mathbb{R}^{K \times D}$ denotes the prototype-enhanced representation that preserves both instance-specific details from the raw input and the rich semantic context captured by the prototypes.

Consider a mini-batch of $N$ pairs of enhanced embeddings $\{(\tilde{\bm{z}}_i^{\text{phy}}, \tilde{\bm{z}}_i^{\text{beha}})\}_{i=1}^N$. We organize them into batch matrices $\tilde{\mathbf{Z}}^{\text{phy}} = [\tilde{\bm{z}}_1^{\text{phy}}, \dots, \tilde{\bm{z}}_N^{\text{phy}}]$ and $\tilde{\mathbf{Z}}^{\text{beha}} = [\tilde{\bm{z}}_1^{\text{beha}}, \dots, \tilde{\bm{z}}_N^{\text{beha}}]$. These matrices simultaneously serve as stored embeddings: $\mathbf{U} = \tilde{\mathbf{Z}}^{\text{phy}} = [\bm{u}_1, \dots, \bm{u}_N]$ and $\mathbf{V} = \tilde{\mathbf{Z}}^{\text{beha}} = [\bm{v}_1, \dots, \bm{v}_N]$. Before retrieval, all embeddings are normalized to unit length: $\|\tilde{\bm{z}}_i^{\text{phy}}\| = \|\tilde{\bm{z}}_i^{\text{beha}}\| = \|\bm{u}_i\| = \|\bm{v}_i\| = 1$. Each individual embedding $\tilde{\bm{z}}_i^{\text{phy}}$ and $\tilde{\bm{z}}_i^{\text{beha}}$ acts as state patterns (queries) to retrieve associated features from the stored patterns in $\mathbf{U}$ and $\mathbf{V}$, respectively:
\begin{align}
    \mathbf{U}_{\tilde{\bm{z}}_i^{\text{phy}}} &= \text{Hopfield}(\tilde{\bm{z}}_i^{\text{phy}}, \mathbf{U}, \mathbf{U}, \beta) \\
    \mathbf{U}_{\tilde{\bm{z}}_i^{\text{beha}}} &= \text{Hopfield}(\tilde{\bm{z}}_i^{\text{beha}}, \mathbf{U}, \mathbf{U}, \beta) \\
    \mathbf{V}_{\tilde{\bm{z}}_i^{\text{phy}}} &= \text{Hopfield}(\tilde{\bm{z}}_i^{\text{phy}}, \mathbf{V}, \mathbf{V}, \beta) \\
    \mathbf{V}_{\tilde{\bm{z}}_i^{\text{beha}}} &= \text{Hopfield}(\tilde{\bm{z}}_i^{\text{beha}}, \mathbf{V}, \mathbf{V}, \beta)
\end{align}
where $\beta$ is the scaling parameter of the Hopfield network. Here, $\mathbf{U}_{\tilde{\bm{z}}_i^{\text{phy}}}$ denotes a physiological-retrieved physiological embedding, while $\mathbf{U}_{\tilde{\bm{z}}_i^{\text{beha}}}$ is a behavioral-retrieved physiological embedding. Similarly, $\mathbf{V}_{\tilde{\bm{z}}_i^{\text{phy}}}$ corresponds to a physiological-retrieved behavioral embedding, and $\mathbf{V}_{\tilde{\bm{z}}_i^{\text{beha}}}$ is a behavioral-retrieved behavioral embedding.

Guided by this memory retrieval mechanism, features that exhibit semantic consistency between the query and stored samples are reinforced, while spurious correlations are suppressed. After retrieval, all retrieved embeddings are re-normalized: $\|\mathbf{U}_{\tilde{\bm{z}}_i^{\text{phy}}}\| = \|\mathbf{U}_{\tilde{\bm{z}}_i^{\text{beha}}}\| = \|\mathbf{V}_{\tilde{\bm{z}}_i^{\text{phy}}}\| = \|\mathbf{V}_{\tilde{\bm{z}}_i^{\text{beha}}}\| = 1$. Drawing inspiration from the CLOOB objective \cite{furst2022cloob}, we define the Semantic‑Level Leave One Out Boost (SemLOOB) loss as:
\begin{equation}
\begin{split}
    \mathcal{L}_{\mathrm{SemLOOB}} &= - \frac{1}{N} \sum_{i=1}^{N} \ln \frac{\exp(\tau_{\text{pcl}}^{-1} \mathbf{U}_{\tilde{\bm{z}}_i^{\text{phy}}}^T \mathbf{U}_{\tilde{\bm{z}}_i^{\text{beha}}})}{\sum_{j \neq i}^{N} \exp(\tau_{\text{pcl}}^{-1} \mathbf{U}_{\tilde{\bm{z}}_i^{\text{phy}}}^T \mathbf{U}_{\tilde{\bm{z}}_j^{\text{beha}}})} \\
    &\quad - \frac{1}{N} \sum_{i=1}^{N} \ln \frac{\exp(\tau_{\text{pcl}}^{-1} \mathbf{V}_{\tilde{\bm{z}}_i^{\text{phy}}}^T \mathbf{V}_{\tilde{\bm{z}}_i^{\text{beha}}})}{\sum_{j \neq i}^{N} \exp(\tau_{\text{pcl}}^{-1} \mathbf{V}_{\tilde{\bm{z}}_j^{\text{phy}}}^T \mathbf{V}_{\tilde{\bm{z}}_i^{\text{beha}}})}
\end{split}
\end{equation}
where $\tau_{\text{pcl}}$ is a temperature parameter that scales the logits.

\subsection{Hierarchical Semantic Compression and Distribution Prediction}
\label{sec:Hierarchical_Semantic_Compression}
After amplifying semantic-level co-occurrences, we concatenate the prototype-enhanced physiological and behavioral features to obtain the fused representation $\tilde{\bm{z}}_{\text{fuse}}\in \mathbb{R}^{2K \times D}$:
\begin{equation}
\tilde{\bm{z}}_{\text{fuse}} = \text{Concat}([\tilde{\bm{z}}^{\text{phy}}, \tilde{\bm{z}}^{\text{beha}}])
\end{equation}

We then introduce a Hierarchical Semantic Compression (HSC) strategy that iteratively abstracts high-level affective concepts from the fine-grained prototype space. Specifically, we stack $L$ compression blocks, each containing a prototype abstraction layer followed by a multi-head self-attention layer (MHSA). Let $\mathbf{H}^{(0)} = \bm{z}_{\text{fuse}} \in \mathbb{R}^{2K \times D}$ denote the initial state. The update for the $l$-th layer ($1 \le l \le L$) is:
\begin{align}
    \tilde{\mathbf{H}}^{(l)} &= \text{Hopfield}(\mathbf{H}^{(l-1)}, \mathbf{W}_{\text{lookup}}^{(l)}, \mathbf{W}_{\text{content}}^{(l)}, \beta) + \mathbf{H}^{(l-1)} \\
    \mathbf{H}^{(l)} &= \text{MHSA}(\tilde{\mathbf{H}}^{(l)}) + \tilde{\mathbf{H}}^{(l)}
\end{align}
where $\tilde{\mathbf{H}}^{(l)}$ is the hidden state of the $l$-th layer. At each layer, the learnable content matrix $\mathbf{W}_{\text{content}}^{(l)} \in \mathbb{R}^{M_l \times D}$ serves as a set of semantic slots with variable capacity $M_l$. Specifically, the slots of the first layer $\mathbf{W}_{\text{content}}^{(1)}$ are initialized from the prototype memory bank ($\mathbf{W}_{\text{memory}} \in \mathbb{R}^{M \times D}$) (Section~\ref{sec:EmotionMemoryBank}), and the slot capacity $M_l$ is progressively reduced across layers. This design gradually compresses fine‑grained prototypes into highly abstract semantic representations whose structure aligns with the target emotion categories.

The output of the final block is mapped into the emotion probability space to obtain the predicted distribution $\hat{\bm{y}}_i \in \mathbb{R}^{E}$, where $E$ is the number of emotion classes. We use KL divergence as the task loss to measure the discrepancy between the prediction $\hat{\bm{y}}_i$ and the ground truth distribution $\bm{y}_i$:
\begin{equation}
\mathcal{L}_{\text{task}} = \frac{1}{N} \sum_{i=1}^{N} D_{\text{KL}}(\bm{y}_i \| \hat{\bm{y}}_i)
\end{equation}
The overall training objective combines the three loss terms:
\begin{equation}
    \mathcal{L}_{\mathrm{total}} = \mathcal{L}_{\mathrm{task}} + \mathcal{L}_{\mathrm{PRD}}+ \mathcal{L}_{\mathrm{SemLOOB}}
\end{equation}
where $\mathcal{L}_{\mathrm{task}}$ aligns predictions with ground truth emotion distribution, $\mathcal{L}_{\mathrm{PRD}}$ enforces cross-modal semantic structural consistency, and $\mathcal{L}_{\mathrm{SemLOOB}}$ promotes cross-modal semantic co-occurrence consistency.

\section{Experiments}

\subsection{Datasets}
To the best of our knowledge, only two public multimodal datasets contain both  physiological and behavioral signals available for emotion distribution learning: DMER \cite{yang2024multimodal} and WESAD \cite{schmidt2018introducing}. We evaluate our MPCL framework on both.

\subsubsection{DMER}
This dataset contains multimodal recording from 80 participants across four modalities: EEG, GSR, PPG, and frontal facial videos. Each participant watched 32 video clips and subsequently rated their emotional states using the 10-item short form of the Positive and Negative Affect Schedule (PANAS) \cite{mackinnon1999short}. The PANAS scale includes five positive emotions (inspired, alert, excited, enthusiastic, determined) and five negative emotions (afraid, upset, nervous, scared, distressed), each rated on a 5‑point Likert scale from 1 (very slightly or not at all) to 5 (extremely). The raw scores are normalized to form emotion distributions, which serve as the ground-truth emotion labels. Due to incomplete physiological recordings for 7 participants, our experiments are conducted on the remaining 73 subjects. 

\subsubsection{WESAD}
This dataset includes multimodal signals collected from 17 participants using wrist- and chest-worn sensors. The recorded modalities include ECG, EMG, EDA, and 3‑axis accelerometer (ACC). Participants underwent stimulus‑based inductions designed to elicit four emotional states: neutral, stress, amusement, and meditation. After each induction, the PANAS scale was used to obtain ground-truth emotion labels. Data from two subjects were excluded because of sensor failures, leaving 15 participants for our experiments. 

\subsection{Data Processing and Feature Extraction}
\label{subsec:data-proc}
To ensure a fair comparison, we follow the preprocessing and feature extraction protocols described in the original dataset publications.

\subsubsection{DMER} EEG signals were filtered using a 1–50Hz band-pass filter and a 50Hz notch filter to remove noise. Artifact components were subsequently eliminated using independent component analysis (ICA) \cite{lee1998independent}. Differential entropy (DE) features were extracted via Short-Time Fourier Transform (STFT) over five frequency bands: $\delta$ (1–3Hz), $\theta$ (4–7Hz), $\alpha$ (8–13Hz), $\beta$ (14–30Hz), and $\gamma$ (31–50Hz). GSR signals were band-pass filtered with a lower cutoff of 0.01Hz and an upper cutoff of 49Hz, while PPG signals were filtered in the 0.01–1.9Hz range. For facial video recordings, Local Binary Patterns on Three-Orthogonal Planes (LBP-TOP) features were computed on each 1-second video segment. In total, we obtained 90 EEG features, 28 GSR features, 27 PPG features, and 768 video features per sample.

\subsubsection{WESAD}
ECG signals were processed with a peak-detection algorithm to identify heartbeats, from which heart rate (HR) and heart rate variability (HRV) features were derived. For EDA, we computed the mean and standard deviation of the signal, then decomposed it into skin conductance level (SCL) and skin conductance response (SCR) components; the number of SCR peaks and their mean amplitude were recorded. For EMG, the mean, standard deviation, range, integral, median, peak frequency and power spectral density (PSD) of the signal were computed. Additionally, the number of peaks and the mean, standard deviation, sum, and normalized sum of peak amplitudes were extracted. For ACC data, the mean, standard deviation, and absolute integral per axis were computed, as well as the magnitude (and sum) across all axes, together with the peak frequency for each axis. Overall, each sample yields 73 ECG features, 4 EDA features, 14 EMG features, and 12 ACC features.

\subsection{Evaluation Metrics and Protocols}

To ensure comparability with previous emotion distribution learning research \cite{liu2023emotion,zheng2025helo}, we evaluate MPCL using six standard metrics. These include four distance measures: Chebyshev ($\downarrow$), Clark ($\downarrow$), Canberra ($\downarrow$), and Kullback-Leibler (KL) divergence ($\downarrow$), and two similarity coefficients: Cosine ($\uparrow$) and Intersection ($\uparrow$). An Average Rank is also reported to summarize overall performance across all six metrics.

To comprehensively evaluate our model on the DMER and WESAD datasets, we adopt both {\it subject-dependent} and {\it subject-independent} settings. In the subject-dependent setting, each subject’s data are randomly split into 80\% for training and 20\% for testing, and the final performance is averaged over all subjects. For the subject-independent evaluation, we employ a Leave-One-Subject-Out (LOSO) scheme, where one subject is held out for testing in each fold while the remaining subjects form the training set. The reported results are averaged over all folds.

\begin{table*}[t]
\centering
\caption{Quantitative comparison with baseline methods on the DMER and WESAD datasets under the subject-dependent setting.}
\label{tab:comparison of subject-dependent}
\resizebox{\textwidth}{!}{%
\begin{tabular}{l c c c c c c c c c c c >{\columncolor{aliceblue}}c}
\toprule
\multirow{2}{*}{Dataset} & \multirow{2}{*}{Mesure} & \multicolumn{3}{c}{LDL} & \multicolumn{1}{c}{MER} & \multicolumn{3}{c}{Single-modal EDL} & \multicolumn{1}{c}{MMER} & \multicolumn{3}{c}{Multimodal EDL} \\
\cmidrule(lr){3-5} \cmidrule(lr){6-6} \cmidrule(lr){7-9} \cmidrule(lr){10-10} \cmidrule(lr){11-13}
 &  & PT-SVM \cite{geng2016label} & AA-KNN \cite{geng2016label} & SA-CPNN \cite{geng2013facial} & MAET \cite{jiang2023multimodal} & LDL-LRR \cite{jia2021label} & TLR-DL \cite{kou2024exploiting} & CAD \cite{wen2023ordinal} & CARAT \cite{peng2024carat} & EmotionDict \cite{liu2023emotion} & HeLo \cite{zheng2025helo} & \textbf{MPCL} \\
\midrule
\multirow{8}{*}{DMER} 
 & Chebyshev ($\downarrow$) & 0.0851 (7) & 0.0949 (10) & 0.0917 (8) & 0.0522 (3) & 0.0936 (9) & 0.0847 (6) & 0.1017 (11) & 0.0752 (4) & 0.0753 (5) & 0.0446 (2) & \textbf{0.0346 (1)} \\
 & Clark ($\downarrow$) & 0.5787 (6) & 0.6649 (10) & 0.6333 (8) & 0.3762 (3) & 0.6529 (9) & 0.5380 (4) & 0.7517 (11) & 0.5951 (7) & 0.5463 (5) & 0.3256 (2) & \textbf{0.2369 (1)} \\
 & Canberra ($\downarrow$) & 1.5294 (7) & 1.2357 (3) & 1.8092 (9) & 0.9875 (3) & 1.8224 (10) & 1.3271 (5) & 1.9415 (11) & 1.5965 (8) & 1.4771 (6) & 0.8664 (2) & \textbf{0.6068 (1)} \\
 & KL ($\downarrow$) & 0.0837 (6) & 0.1409 (10) & 0.1048 (8) & 0.0288 (2) & 0.0644 (4) & 0.0925 (7) & 0.1460 (11) & 0.1137 (9) & 0.0766 (5) & 0.0323 (3) & \textbf{0.0177 (1)} \\
 & Cosine ($\uparrow$) & 0.9237 (8) & 0.9315 (6) & 0.9058 (9) & 0.9688 (3) & 0.9016 (10) & 0.9339 (4) & 0.8923 (11) & 0.9332 (5) & 0.9313 (7) & 0.9714 (2) & \textbf{0.9874 (1)} \\
 & Intersection ($\uparrow$) & 0.8417 (8) & 0.8425 (7) & 0.8150 (9) & 0.9113 (3) & 0.8136 (10) & 0.8537 (5) & 0.8077 (11) & 0.8474 (6) & 0.8492 (5) & 0.9128 (2) & \textbf{0.9427 (1)} \\
 \cmidrule{2-13}
 & Average Rank ($\downarrow$) & 7.00 (7) & 7.67 (8) & 8.50 (9) & 2.83 (3) & 8.67 (10) & 5.17 (4) & 11.00 (11) & 6.50 (6) & 5.50 (5) & 2.17 (2) & \textbf{1.00 (1)} \\
\midrule
\multirow{8}{*}{WESAD} 
 & Chebyshev ($\downarrow$) & 0.0515 (8) & 0.0346 (4) & 0.0998 (11) & 0.0347 (5) & 0.0357 (7) & 0.0622 (10) & 0.0569 (9) & 0.0354 (6) & 0.0090 (3) & 0.0073 (2) & \textbf{0.0046 (1)} \\
 & Clark ($\downarrow$) & 0.3940 (8) & 0.2779 (4) & 0.7285 (11) & 0.3503 (7) & 0.2822 (6) & 0.5381 (10) & 0.4618 (9) & 0.2812 (5) & 0.0733 (3) & 0.0653 (2) & \textbf{0.0417 (1)} \\
 & Canberra ($\downarrow$) & 1.0249 (8) & 0.7381 (4) & 2.0746 (11) & 0.9100 (7) & 0.7466 (6) & 1.3586 (10) & 1.2199 (9) & 0.7430 (5) & 0.1876 (3) & 0.1614 (2) & \textbf{0.1026 (1)} \\
 & KL ($\downarrow$) & 0.0320 (8) & 0.0243 (5) & 0.1226 (10) & 0.0251 (6) & 0.0273 (7) & 0.2051 (11) & 0.0440 (9) & 0.0210 (4) & 0.0017 (3) & 0.0010 (2) & \textbf{0.0003 (1)} \\
 & Cosine ($\uparrow$) & 0.9728 (8) & 0.9793 (7) & 0.8918 (11) & 0.9867 (4) & 0.9821 (6) & 0.9541 (10) & 0.9624 (9) & 0.9824 (5) & 0.9985 (3) & 0.9992 (2) & \textbf{0.9998 (1)} \\
 & Intersection ($\uparrow$) & 0.9024 (8) & 0.9297 (4) & 0.7896 (11) & 0.9213 (7) & 0.9292 (6) & 0.8746 (10) & 0.8812 (9) & 0.9296 (5) & 0.9820 (3) & 0.9905 (2) & \textbf{0.9974 (1)} \\
 \cmidrule{2-13}
 & Average Rank ($\downarrow$) & 8.00 (8) & 4.67 (4) & 10.83 (11) & 6.00 (6) & 6.33 (7) & 10.17 (10) & 9.00 (9) & 5.00 (5) & 3.00 (3) & 2.00 (2) & \textbf{1.00 (1)} \\
\bottomrule
\end{tabular}%
}
\\[5pt]
\parbox{\textwidth}{\footnotesize
Arrows $\downarrow$ and $\uparrow$ indicate ``lower is better'' and ``higher is better'', respectively. Parenthesized values denote per‑metric rankings and the average rank. The best result is highlighted in bold.
}
\end{table*}

\subsection{Implementation Details}
Our method is implemented in PyTorch and trained on an NVIDIA GeForce RTX 4090 GPU. We use the Adam optimizer with a learning rate of 0.001, a batch size of 128, and train for 400 epochs. For the DMER dataset, EEG is treated as the primary modality,  with GSR and PPG as auxiliary physiological modalities; facial videos constitute the behavioral signal. For the WESAD dataset, ECG serves as the primary modality, while EDA and EMG are auxiliary modalities; ACC data is used as the behavioral signal. In the MSAF module, embedding dimension $D=128$, and the scaling set $\mathcal{B} = \{\beta_{\text{low}}, \beta_{\text{mid}}, \beta_{\text{high}}\}$ is $\{8, 14.3,22\}$. For the prototype memory bank, the number of prototypes $M=100$, and the temperature parameter $\tau_{\text{dist}}=0.07$. For the PCL module, the scaling $\beta=14.3$, and the inverse temperature parameter $\tau_{\text{pcl}}^{-1}=50$. In the HSC module, the number of compression blocks $L=10$, and the capacity $M_l$ of the content matrix is progressively reduced from 100 in the first layer to the number of emotion classes in the last layer.

\begin{figure*}[!t]
\centering
\includegraphics[width=\linewidth]{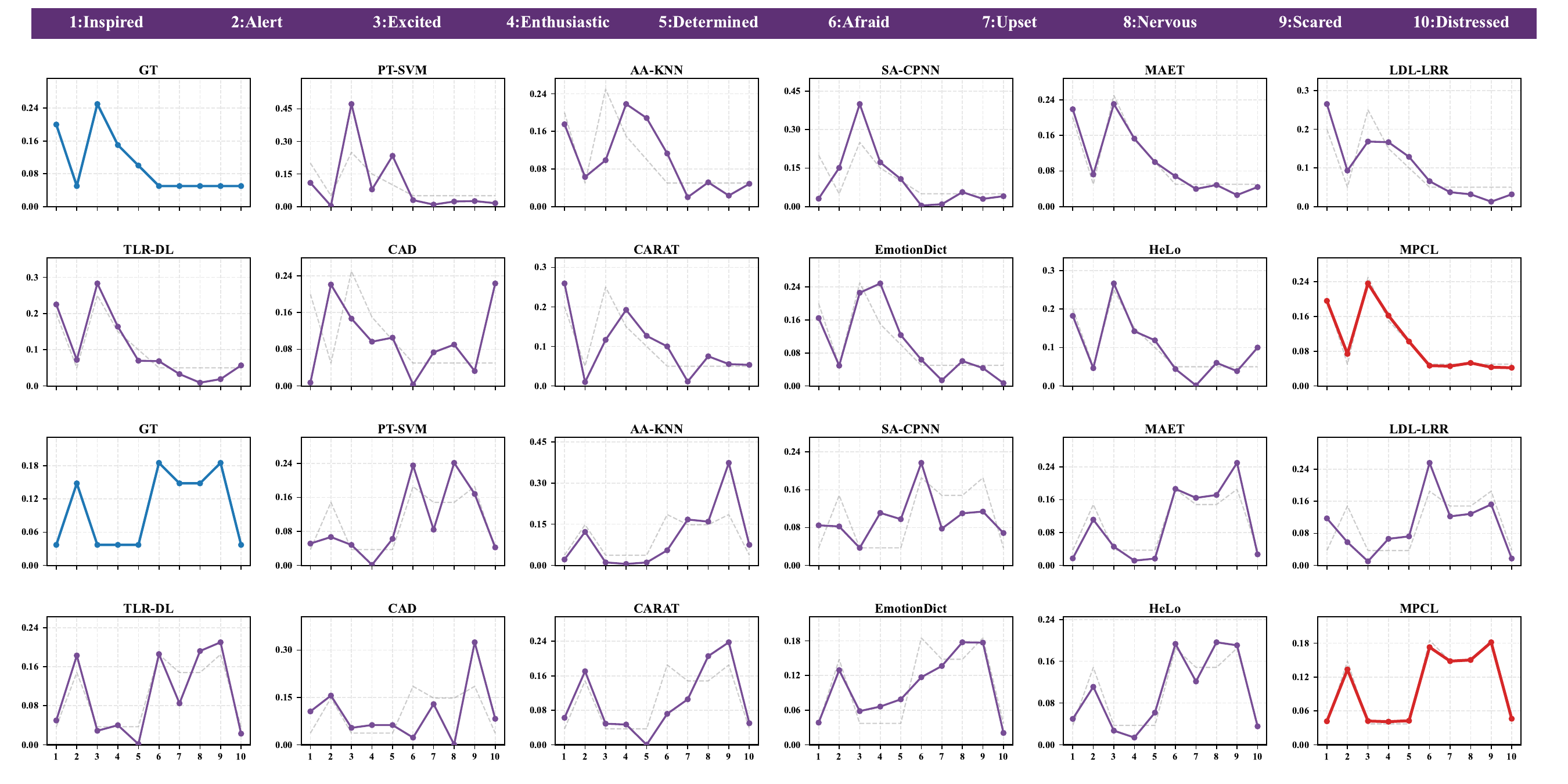}
\caption{Comparison of emotion distribution predictions between MPCL and state-of-the-art baselines on two representative samples (Subject 26). GT denotes the ground-truth distribution. Emotion indices 1–10 correspond to: inspired, alert, excited, enthusiastic, determined, afraid, upset, nervous, scared, and distressed.}
\label{fig_compare}
\end{figure*}

\begin{table*}[t]
\centering
\caption{Quantitative comparison with baseline methods on the DMER and WESAD datasets under the subject-independent setting.}
\label{tab:comparison_subject_independent}
\resizebox{\textwidth}{!}{%
\begin{tabular}{l c c c c c c c c c c c >{\columncolor{aliceblue}}c}
\toprule
\multirow{2}{*}{Dataset} & \multirow{2}{*}{Measure} & \multicolumn{3}{c}{LDL} & \multicolumn{1}{c}{MER} & \multicolumn{3}{c}{Single-modal EDL} & \multicolumn{1}{c}{MMER} & \multicolumn{3}{c}{Multimodal EDL} \\
\cmidrule(lr){3-5} \cmidrule(lr){6-6} \cmidrule(lr){7-9} \cmidrule(lr){10-10} \cmidrule(lr){11-13}
 &  & PT-SVM \cite{geng2016label} & AA-KNN \cite{geng2016label} & SA-CPNN \cite{geng2013facial} & MAET \cite{jiang2023multimodal} & LDL-LRR \cite{jia2021label} & TLR-DL \cite{kou2024exploiting} & CAD \cite{wen2023ordinal} & CARAT \cite{peng2024carat} & EmotionDict \cite{liu2023emotion} & HeLo \cite{zheng2025helo} & \textbf{MPCL} \\
\midrule
\multirow{8}{*}{DMER} 
 & Chebyshev ($\downarrow$) & 0.1462 (11) & 0.1070 (10) & 0.0928 (4) & 0.0939 (5) & 0.1012 (8) & 0.0964 (7) & 0.0925 (3) & 0.1023 (9) & 0.0945 (6) & 0.0882 (2) & \textbf{0.0763 (1)} \\
 & Clark ($\downarrow$) & 1.2877 (11) & 0.6860 (9) & 0.6440 (4) & 0.6605 (5) & 0.6856 (8) & 0.6702 (7) & 0.6379 (3) & 0.6909 (10) & 0.6676 (6) & 0.6289 (2) & \textbf{0.5932 (1)} \\
 & Canberra ($\downarrow$) & 2.0459 (11) & 1.8308 (7) & 1.8542 (8) & 1.7863 (3) & 1.9071 (9) & 1.8244 (6) & 1.8104 (5) & 1.9161 (10) & 1.8092 (4) & 1.7603 (2) & \textbf{1.5843 (1)} \\
 & KL ($\downarrow$) & 0.5528 (11) & 0.1227 (9) & 0.1089 (5) & 0.1088 (4) & 0.1222 (8) & 0.1142 (7) & 0.1076 (3) & 0.1245 (10) & 0.1104 (6) & 0.1027 (2) & \textbf{0.1009 (1)} \\
 & Cosine ($\uparrow$) & 0.8295 (11) & 0.8843 (10) & 0.9022 (4) & 0.9016 (5) & 0.8917 (8) & 0.8964 (7) & 0.9036 (3) & 0.8899 (9) & 0.9001 (6) & 0.9148 (2) & \textbf{0.9273 (1)} \\
 & Intersection ($\uparrow$) & 0.7762 (11) & 0.8050 (8) & 0.8102 (6) & 0.8147 (3) & 0.8037 (9) & 0.8087 (7) & 0.8122 (4) & 0.8026 (10) & 0.8119 (5) & 0.8194 (2) & \textbf{0.8496 (1)} \\
 \cmidrule{2-13}
 & Average Rank ($\downarrow$) & 11.00 (11) & 8.83 (9) & 5.17 (5) & 4.17 (4) & 8.33 (8) & 6.83 (7) & 3.50 (3) & 9.67 (10) & 5.50 (6) & 2.00 (2) & \textbf{1.00 (1)} \\
\midrule
\multirow{8}{*}{WESAD} 
 & Chebyshev ($\downarrow$) & 0.0471 (7) & 0.0500 (9) & 0.0894 (11) & 0.0516 (10) & 0.0466 (6) & 0.0441 (4) & 0.0481 (8) & 0.0454 (5) & 0.0421 (3) & 0.0403 (2) & \textbf{0.0314 (1)} \\
 & Clark ($\downarrow$) & 0.4506 (8) & 0.4516 (10) & 0.6394 (11) & 0.4514 (9) & 0.4471 (6) & 0.4145 (4) & 0.4488 (7) & 0.4270 (5) & 0.3631 (3) & 0.3455 (2) & \textbf{0.2973 (1)} \\
 & Canberra ($\downarrow$) & 1.1928 (10) & 1.1762 (7) & 1.7384 (11) & 1.1666 (6) & 1.1817 (8) & 1.0746 (4) & 1.1647 (5) & 1.1874 (9) & 0.9530 (3) & 0.9329 (2) & \textbf{0.8685 (1)} \\
 & KL ($\downarrow$) & 0.0347 (6) & 0.0423 (9) & 0.0944 (11) & 0.0445 (10) & 0.0343 (4) & 0.0372 (8) & 0.0366 (7) & 0.0345 (5) & 0.0282 (2) & 0.0283 (3) & \textbf{0.0228 (1)} \\
 & Cosine ($\uparrow$) & 0.9727 (6) & 0.9663 (10) & 0.9154 (11) & 0.9678 (9) & 0.9731 (4) & 0.9695 (8) & 0.9711 (7) & 0.9731 (4) & 0.9773 (3) & 0.9790 (2) & \textbf{0.9857 (1)} \\
 & Intersection ($\uparrow$) & 0.8935 (7) & 0.8933 (9) & 0.8254 (11) & 0.8967 (4) & 0.8945 (5) & 0.8917 (10) & 0.8935 (8) & 0.8944 (6) & 0.9126 (3) & 0.9154 (2) & \textbf{0.9371 (1)} \\
 \cmidrule{2-13}
 & Average Rank ($\downarrow$) & 7.33 (8) & 9.00 (10) & 11.00 (11) & 8.00 (9) & 5.50 (4) & 6.33 (6) & 7.00 (7) & 5.67 (5) & 2.83 (3) & 2.17 (2) & \textbf{1.00 (1)} \\
\bottomrule
\end{tabular}%
}
\\[5pt]
\parbox{\textwidth}{\footnotesize
Arrows $\downarrow$ and $\uparrow$ indicate ``lower is better'' and ``higher is better'', respectively. Parenthesized values denote per‑metric rankings and the average rank. The best result is highlighted in bold.
}
\end{table*}

\subsection{Experimental Results}

\subsubsection{Baselines} 
To evaluate the effectiveness of MPCL, we compare it against a comprehensive set of representative baselines on both the DMER and WESAD datasets. These baselines span five categories: 
\begin{itemize}
    \item Classical label distribution learning (LDL): PT-SVM \cite{geng2016label}, AA-KNN \cite{geng2016label}, SA-CPNN \cite{geng2013facial}.
    \item Multimodal emotion recognition (MER): MAET \cite{jiang2023multimodal}.
    \item Single-modal emotion distribution learning (single-modal EDL): LDL-LRR \cite{jia2021label}, TLRLDL \cite{kou2024exploiting}, CAD \cite{wen2023ordinal}.
    \item Multimodal multi-label emotion recognition (MMER): CARAT \cite{peng2024carat}.
    \item Recent multimodal EDL (Multimodal EDL): EmotionDict \cite{liu2023emotion}, Helo \cite{zheng2025helo}).
\end{itemize}
As MER and MMER methods are originally designed for categorical classification, we apply a softmax function to their output layer to generate probability distributions.

\subsubsection{Subject-dependent Experiments}

Table~\ref{tab:comparison of subject-dependent} reports the quantitative results under the subject-dependent setting. MPCL consistently outperforms the all baselines across all six evaluation metrics on both datasets. In general, deep-learning-based approaches (MER, single-modal LDL, MMLER, and multimodal EDL) surpass classical LDL methods. 

We further visualize the ground-truth and predicted emotion distributions for selected samples from the DMER dataset (Fig.~\ref{fig_compare}). The emotion distributions exhibit clear semantic dependencies, reflected in the co-occurrence of emotions sharing the same valence. For instance, the first row shows simultaneous activation of positive emotions (including inspired, excited, enthusiastic, determined), while the second row presents a cluster of negative emotions (afraid, upset, nervous, scared). Most baselines fail to capture these intrinsic correlations and produce predictions with noticeable fluctuations. In contrast, MPCL effectively models the distribution patterns of co-occurring emotions, yielding predictions that align closely with the ground truth.

\subsubsection{Subject-independent Experiments}

Table \ref{tab:comparison_subject_independent} presents the results under the more challenging subject‑independent setting, where inter‑subject variability—especially in EEG signals—poses significant difficulties. MPCL again achieves the best performance on both datasets, demonstrating its ability to extract stable emotion‑related features across different individuals.

Overall, by enhancing semantic‑level co‑occurrence among emotion prototypes, MPCL effectively captures the global topological structure of emotions. Both quantitative and qualitative experiments confirm the superior predictive capability of our method.

\subsection{Ablation Study}

\begin{table}[t]
\centering
\caption{Ablation study of different components on the DMER dataset. ``\textit{w/o}'' indicates the removal of the corresponding module.}
\label{tab:ablation of dmer}
\resizebox{\linewidth}{!}{%
\begin{tabular}{l l c c c c c}
\toprule
\multirow{2}{*}{Settings} & \multirow{2}{*}{Measure} & \textit{w/o} & \textit{w/o} & \textit{w/o} & \textit{w/o} & \cellcolor{aliceblue} \\
 & & MSAF & PRD & PCL & HSC & \cellcolor{aliceblue}\multirow{-2}{*}{\textbf{MPCL}} \\
\midrule
\multirow{6}{*}{\shortstack{Subject-\\Dependent}} 
 & Chebyshev ($\downarrow$)   & 0.0385 & 0.0425 & 0.0396 & 0.0388 & \cellcolor{aliceblue}\textbf{0.0346} \\
 & Clark ($\downarrow$)       & 0.2643 & 0.2952 & 0.2914 & 0.2704 & \cellcolor{aliceblue}\textbf{0.2369} \\
 & Canberra ($\downarrow$)    & 0.6652 & 0.7759 & 0.7466 & 0.6861 & \cellcolor{aliceblue}\textbf{0.6068} \\
 & KL ($\downarrow$)          & 0.0258 & 0.0332 & 0.0317 & 0.0284 & \cellcolor{aliceblue}\textbf{0.0177} \\
 & Cosine ($\uparrow$)        & 0.9766 & 0.9714 & 0.9702 & 0.9743 & \cellcolor{aliceblue}\textbf{0.9874} \\
 & Intersection ($\uparrow$)  & 0.9326 & 0.9186 & 0.9221 & 0.9297 & \cellcolor{aliceblue}\textbf{0.9427} \\
\midrule
\multirow{6}{*}{\shortstack{Subject-\\Independent}} 
 & Chebyshev ($\downarrow$)   & 0.0804 & 0.0885 & 0.0871 & 0.0824 & \cellcolor{aliceblue}\textbf{0.0763} \\
 & Clark ($\downarrow$)       & 0.6253 & 0.6843 & 0.6759 & 0.6381 & \cellcolor{aliceblue}\textbf{0.5932} \\
 & Canberra ($\downarrow$)    & 1.6681 & 1.7925 & 1.7638 & 1.6952 & \cellcolor{aliceblue}\textbf{1.5843} \\
 & KL ($\downarrow$)          & 0.1138 & 0.1342 & 0.1296 & 0.1173 & \cellcolor{aliceblue}\textbf{0.1009} \\
 & Cosine ($\uparrow$)        & 0.9145 & 0.8924 & 0.8967 & 0.9085 & \cellcolor{aliceblue}\textbf{0.9273} \\
 & Intersection ($\uparrow$)  & 0.8362 & 0.8125 & 0.8168 & 0.8294 & \cellcolor{aliceblue}\textbf{0.8496} \\
\bottomrule
\end{tabular}%
}
\end{table}

\begin{table}[t]
\centering
\caption{Ablation study of different components on WESAD dataset. ``\textit{w/o}'' indicates the removal of the corresponding module.}
\label{tab:ablation of wesad}
\resizebox{\linewidth}{!}{%
\begin{tabular}{l l c c c c c}
\toprule
\multirow{2}{*}{Settings} & \multirow{2}{*}{Measure} & \textit{w/o} & \textit{w/o} & \textit{w/o} & \textit{w/o} & \cellcolor{aliceblue} \\
 & & MSAF & PRD & PCL & HSC & \cellcolor{aliceblue}\multirow{-2}{*}{\textbf{MPCL}} \\
\midrule
\multirow{6}{*}{\shortstack{Subject-\\Dependent}} 
 & Chebyshev ($\downarrow$)   & 0.0062 & 0.0081 & 0.0083 & 0.0068 & \cellcolor{aliceblue}\textbf{0.0046} \\
 & Clark ($\downarrow$)       & 0.0525 & 0.0692 & 0.0674 & 0.0563 & \cellcolor{aliceblue}\textbf{0.0417} \\
 & Canberra ($\downarrow$)    & 0.1263 & 0.1682 & 0.1625 & 0.1342 & \cellcolor{aliceblue}\textbf{0.1026} \\
 & KL ($\downarrow$)          & 0.0007 & 0.0015 & 0.0014 & 0.0009 & \cellcolor{aliceblue}\textbf{0.0003} \\
 & Cosine ($\uparrow$)        & 0.9984 & 0.9961 & 0.9954 & 0.9980 & \cellcolor{aliceblue}\textbf{0.9998} \\
 & Intersection ($\uparrow$)  & 0.9937 & 0.9842 & 0.9865 & 0.9912 & \cellcolor{aliceblue}\textbf{0.9974} \\
\midrule
\multirow{6}{*}{\shortstack{Subject-\\Independent}} 
 & Chebyshev ($\downarrow$)   & 0.0382 & 0.0486 & 0.0475 & 0.0412 & \cellcolor{aliceblue}\textbf{0.0314} \\
 & Clark ($\downarrow$)       & 0.3247 & 0.3642 & 0.3598 & 0.3325 & \cellcolor{aliceblue}\textbf{0.2973} \\
 & Canberra ($\downarrow$)    & 0.9356 & 1.0542 & 1.0268 & 0.9584 & \cellcolor{aliceblue}\textbf{0.8685} \\
 & KL ($\downarrow$)          & 0.0292 & 0.0385 & 0.0372 & 0.0315 & \cellcolor{aliceblue}\textbf{0.0228} \\
 & Cosine ($\uparrow$)        & 0.9783 & 0.9654 & 0.9682 & 0.9745 & \cellcolor{aliceblue}\textbf{0.9857} \\
 & Intersection ($\uparrow$)  & 0.9211 & 0.8954 & 0.9012 & 0.9156 & \cellcolor{aliceblue}\textbf{0.9371} \\
\bottomrule
\end{tabular}%
}
\end{table}

\subsubsection{Ablation on Main Components}

To validate the effectiveness of each core component in the MPCL framework, we conducted ablation studies on four key modules: multi-scale associative fusion (MSAF), prototype relation distillation (PRD), prototypical co-occurrence learning (PCL), and hierarchical semantic compression (HSC). Tables~\ref{tab:ablation of dmer} and~\ref{tab:ablation of wesad} summarize the results (without each component) on DMER and WESAD under both subject‑dependent and subject‑independent settings. The results indicate that each component contributes significantly to overall performance. Specifically, PRD and PCL are the most influential modules—PRD enforces cross‑modal prototype‑topology alignment, while PCL amplifies semantic‑level co‑occurrence among prototypes. HSC follows as a strong secondary contributor, progressively abstracting fine‑grained prototypes into compact emotion representations. Although MSAF exhibits a relatively moderate impact, it remains essential for effectively fusing multimodal physiological data.

\subsubsection{Ablation on Modalities}

\begin{table}[t]
\centering
\caption{Ablation study of different modalities on DMER. ``\textit{w/o}'' means the removal of the corresponding modality from the input.}
\label{tab:ablation_modalities_DMER}
\resizebox{\linewidth}{!}{%
\begin{tabular}{l l c c c c c}
\toprule
\multirow{2}{*}{Settings} & \multirow{2}{*}{Measure} & \textit{w/o} & \textit{w/o} & \textit{w/o} & \textit{w/o} & \cellcolor{aliceblue} \\
 & & PPG & GSR & EEG & Video & \cellcolor{aliceblue}\multirow{-2}{*}{\textbf{MPCL}} \\
\midrule
\multirow{6}{*}{\shortstack{Subject-\\Dependent}} 
 & Chebyshev ($\downarrow$)   & 0.0362 & 0.0371 & 0.0412 & 0.0465 & \cellcolor{aliceblue}\textbf{0.0346} \\
 & Clark ($\downarrow$)       & 0.2485 & 0.2533 & 0.2847 & 0.3105 & \cellcolor{aliceblue}\textbf{0.2369} \\
 & Canberra ($\downarrow$)    & 0.6250 & 0.6384 & 0.7259 & 0.8214 & \cellcolor{aliceblue}\textbf{0.6068} \\
 & KL ($\downarrow$)          & 0.0198 & 0.0215 & 0.0296 & 0.0368 & \cellcolor{aliceblue}\textbf{0.0177} \\
 & Cosine ($\uparrow$)        & 0.9825 & 0.9804 & 0.9735 & 0.9668 & \cellcolor{aliceblue}\textbf{0.9874} \\
 & Intersection ($\uparrow$)  & 0.9385 & 0.9356 & 0.9242 & 0.9135 & \cellcolor{aliceblue}\textbf{0.9427} \\
\midrule
\multirow{6}{*}{\shortstack{Subject-\\Independent}} 
 & Chebyshev ($\downarrow$)   & 0.0782 & 0.0805 & 0.0894 & 0.0965 & \cellcolor{aliceblue}\textbf{0.0763} \\
 & Clark ($\downarrow$)       & 0.6085 & 0.6174 & 0.6628 & 0.7012 & \cellcolor{aliceblue}\textbf{0.5932} \\
 & Canberra ($\downarrow$)    & 1.6150 & 1.6423 & 1.7456 & 1.8845 & \cellcolor{aliceblue}\textbf{1.5843} \\
 & KL ($\downarrow$)          & 0.1065 & 0.1102 & 0.1287 & 0.1453 & \cellcolor{aliceblue}\textbf{0.1009} \\
 & Cosine ($\uparrow$)        & 0.9215 & 0.9184 & 0.9015 & 0.8842 & \cellcolor{aliceblue}\textbf{0.9273} \\
 & Intersection ($\uparrow$)  & 0.8425 & 0.8386 & 0.8205 & 0.8043 & \cellcolor{aliceblue}\textbf{0.8496} \\
\bottomrule
\end{tabular}%
}
\end{table}

\begin{table}[t]
\centering
\caption{Ablation study of different modalities on WESAD. ``\textit{w/o}'' means the removal of the corresponding modality from the input.}
\label{tab:ablation_modalities_WESAD}
\resizebox{\linewidth}{!}{%
\begin{tabular}{l l c c c c c}
\toprule
\multirow{2}{*}{Settings} & \multirow{2}{*}{Measure} & \textit{w/o} & \textit{w/o} & \textit{w/o} & \textit{w/o} & \cellcolor{aliceblue} \\
 & & EMG & EDA & ECG & ACC & \cellcolor{aliceblue}\multirow{-2}{*}{\textbf{MPCL}} \\
\midrule
\multirow{6}{*}{\shortstack{Subject-\\Dependent}} 
 & Chebyshev ($\downarrow$)   & 0.0052 & 0.0055 & 0.0075 & 0.0092 & \cellcolor{aliceblue}\textbf{0.0046} \\
 & Clark ($\downarrow$)       & 0.0458 & 0.0472 & 0.0585 & 0.0734 & \cellcolor{aliceblue}\textbf{0.0417} \\
 & Canberra ($\downarrow$)    & 0.1105 & 0.1142 & 0.1428 & 0.1856 & \cellcolor{aliceblue}\textbf{0.1026} \\
 & KL ($\downarrow$)          & 0.0004 & 0.0005 & 0.0011 & 0.0019 & \cellcolor{aliceblue}\textbf{0.0003} \\
 & Cosine ($\uparrow$)        & 0.9992 & 0.9990 & 0.9972 & 0.9954 & \cellcolor{aliceblue}\textbf{0.9998} \\
 & Intersection ($\uparrow$)  & 0.9962 & 0.9958 & 0.9885 & 0.9812 & \cellcolor{aliceblue}\textbf{0.9974} \\
\midrule
\multirow{6}{*}{\shortstack{Subject-\\Independent}} 
 & Chebyshev ($\downarrow$)   & 0.0352 & 0.0361 & 0.0425 & 0.0515 & \cellcolor{aliceblue}\textbf{0.0314} \\
 & Clark ($\downarrow$)       & 0.3105 & 0.3142 & 0.3426 & 0.3758 & \cellcolor{aliceblue}\textbf{0.2973} \\
 & Canberra ($\downarrow$)    & 0.8950 & 0.9025 & 0.9854 & 1.0865 & \cellcolor{aliceblue}\textbf{0.8685} \\
 & KL ($\downarrow$)          & 0.0256 & 0.0268 & 0.0345 & 0.0415 & \cellcolor{aliceblue}\textbf{0.0228} \\
 & Cosine ($\uparrow$)        & 0.9815 & 0.9802 & 0.9715 & 0.9604 & \cellcolor{aliceblue}\textbf{0.9857} \\
 & Intersection ($\uparrow$)  & 0.9285 & 0.9264 & 0.9085 & 0.8872 & \cellcolor{aliceblue}\textbf{0.9371} \\
\bottomrule
\end{tabular}%
}
\end{table}

We further evaluate the contribution of each individual modality through ablation studies (Tables~\ref{tab:ablation_modalities_DMER} and~\ref{tab:ablation_modalities_WESAD}). All modalities consistently improve performance under both subject-dependent and subject-independent settings. For physiological signals, the primary modality (EEG on DMER, ECG on WESAD) contributes more substantially than the auxiliary ones (PPG/GSR on DMER, EMG/EDA on WESAD). The behavioral modality also yields significant gains, confirming that the integration of multimodal data is crucial for robust emotion distribution learning.

\subsubsection{Comparison with Alternative Designs}

To demonstrate the advantages of the Hopfield‑based modules in MPCL, we compare them against standard alternative designs. The following configurations are evaluated:
\begin{itemize}
    \item \textbf{MSAF vs. Transformer cross-attention Fusion:} MSAF is replaced with standard Transformer cross‑attention (EEG/ECG as Query, auxiliary physiological signals as Key/Value).

    \item \textbf{Prototype-level vs. Feature-level Co-occurrence:} The prototype-enhanced representations $\tilde{\bm{z}}^{\text{phy}}$ and $\tilde{\bm{z}}^{\text{beha}}$ in PCL are replaced with raw features $\bm{z}^{\text{phy}}$ and $\bm{z}^{\text{beha}}$. This modification degrades prototype-level co-occurrence to feature-level co-occurrence.

    \item \textbf{Memory-guided Retrieval vs. InfoNCE Alignment:} The memory‑retrieval mechanism is removed and the SemLOOB loss is replaced by InfoNCE loss (adapted from CLIP), i.e., directly performing cross-modal alignment on the prototype-enhanced representations $\tilde{\bm{z}}^{\text{phy}}$ and $\tilde{\bm{z}}^{\text{beha}}$.

    \item \textbf{HSC vs. Transformer Encoder:} HSC is replaced with standard Transformer encoder layers while keeping the same number of layers $L$.
\end{itemize}

Quantitative results (Table~\ref{tab:alternative_designs}) show the following:
\begin{itemize}
    \item MSAF outperforms Transformer cross‑attention fusion, benefiting from its energy‑minimization formulation and multi‑scale associative memory retrieval. Unlike standard attention mechanisms, Hopfield networks allow flexible tuning of the inverse temperature $\beta$, since the fixed-point dynamics is not strictly coupled to the dimension of the associative space $d_k$. Consequently, this flexibility enables MSAF to effectively capture complementary information across heterogeneous signals at varying granularities.
    \item Prototype-level co-occurrence surpasses feature‑level co-occurrence, confirming that prototypes effectively disentangle semantics and suppress noise, and amplifying co-occurrence within this space effectively captures more meaningful emotion representations.
    \item Memory‑guided retrieval yields better results than global contrastive alignment, indicating that the Hopfield retrieval process effectively amplifies features exhibiting semantic consistency between input and stored patterns while suppressing spurious correlations, thereby revealing latent structural associations among emotions at a semantic level.
    \item HSC is more effective than a plain Transformer encoder in simulating the cognitive progression from concrete signals to abstract affective concepts. Its bottom‑up compression strategy—progressively reducing semantic‑slot capacity ($M_l$) across layers—forces the model to condense fine‑grained prototypes into highly abstract representations, thereby enabling accurate emotion distribution prediction.
\end{itemize}

\begin{table*}[t]
\centering
\scriptsize
\setlength{\tabcolsep}{2.5pt}
\renewcommand{\arraystretch}{1.15}
\caption{Comparison with alternative designs on DMER and WESAD. ``\textit{w/}'' indicates that the corresponding MPCL module is replaced by the specified alternative: CrossAttn (replacing MSAF), Feat-Co (replacing prototype-level co-occurrence), InfoNCE (replacing memory-guided retrieval), and TransEnc (replacing HSC).}
\vspace{-6pt}
\label{tab:alternative_designs}

\begin{tabular*}{\linewidth}{@{\extracolsep{\fill}} c l ccccc ccccc}
\toprule
\multirow{2.5}{*}{\textbf{Protocol}} & \multirow{2.5}{*}{\textbf{Measure}} & \multicolumn{5}{c}{\textbf{DMER Dataset}} & \multicolumn{5}{c}{\textbf{WESAD Dataset}} \\
\cmidrule(lr){3-7} \cmidrule(lr){8-12}
 & & w/ CrossAttn & w/ Feat-Co & w/ InfoNCE & w/ TransEnc & \textbf{MPCL} & w/ CrossAttn & w/ Feat-Co & w/ InfoNCE & w/ TransEnc & \textbf{MPCL} \\
\midrule
\multirow{6}{*}{\shortstack{Subject-\\Dependent}} 
 & Chebyshev ($\downarrow$)   & 0.0363 & 0.0382 & 0.0374 & 0.0368 & \textbf{0.0346} & 0.0054 & 0.0067 & 0.0064 & 0.0059 & \textbf{0.0046} \\
 & Clark ($\downarrow$)       & 0.2497 & 0.2743 & 0.2648 & 0.2541 & \textbf{0.2369} & 0.0483 & 0.0556 & 0.0537 & 0.0498 & \textbf{0.0417} \\
 & Canberra ($\downarrow$)    & 0.6358 & 0.7012 & 0.6784 & 0.6486 & \textbf{0.6068} & 0.1147 & 0.1384 & 0.1319 & 0.1217 & \textbf{0.1026} \\
 & KL ($\downarrow$)          & 0.0213 & 0.0287 & 0.0246 & 0.0238 & \textbf{0.0177} & 0.0006 & 0.0011 & 0.0009 & 0.0007 & \textbf{0.0003} \\
 & Cosine ($\uparrow$)        & 0.9823 & 0.9748 & 0.9782 & 0.9791 & \textbf{0.9874} & 0.9989 & 0.9972 & 0.9976 & 0.9984 & \textbf{0.9998} \\
 & Intersection ($\uparrow$)  & 0.9367 & 0.9284 & 0.9316 & 0.9352 & \textbf{0.9427} & 0.9961 & 0.9918 & 0.9924 & 0.9953 & \textbf{0.9974} \\
\midrule
\multirow{6}{*}{\shortstack{Subject-\\Independent}} 
 & Chebyshev ($\downarrow$)   & 0.0794 & 0.0846 & 0.0817 & 0.0803 & \textbf{0.0763} & 0.0346 & 0.0414 & 0.0378 & 0.0367 & \textbf{0.0314} \\
 & Clark ($\downarrow$)       & 0.6124 & 0.6542 & 0.6389 & 0.6217 & \textbf{0.5932} & 0.3118 & 0.3456 & 0.3294 & 0.3207 & \textbf{0.2973} \\
 & Canberra ($\downarrow$)    & 1.6356 & 1.7248 & 1.6892 & 1.6583 & \textbf{1.5843} & 0.8943 & 0.9654 & 0.9287 & 0.9248 & \textbf{0.8685} \\
 & KL ($\downarrow$)          & 0.1087 & 0.1243 & 0.1164 & 0.1126 & \textbf{0.1009} & 0.0264 & 0.0326 & 0.0298 & 0.0282 & \textbf{0.0228} \\
 & Cosine ($\uparrow$)        & 0.9214 & 0.9056 & 0.9123 & 0.9158 & \textbf{0.9273} & 0.9813 & 0.9718 & 0.9746 & 0.9782 & \textbf{0.9857} \\
 & Intersection ($\uparrow$)  & 0.8423 & 0.8264 & 0.8327 & 0.8356 & \textbf{0.8496} & 0.9294 & 0.9153 & 0.9196 & 0.9224 & \textbf{0.9371} \\
\bottomrule
\end{tabular*}
\end{table*}

\subsubsection{Hyperparameter Analysis}

\begin{table*}[t]
    \centering
    \setlength{\tabcolsep}{2.5pt} 
    \caption{Ablation study of different scaling strategies ($\beta$) in the MSAF module on the DMER dataset under both Subject-Dependent and Subject-Independent protocols.}
    \vspace{-6pt}
    \label{tab:ablation_beta_dmer}
    \resizebox{\textwidth}{!}{
        \begin{tabular}{c|cccccc|cccccc}
            \toprule
            \multirow{2}{*}{\textbf{Configuration}} & \multicolumn{6}{c|}{\textbf{Subject-Dependent}} & \multicolumn{6}{c}{\textbf{Subject-Independent}} \\
            \cmidrule(lr){2-7} \cmidrule(lr){8-13}
             & Chebyshev ($\downarrow$) & Clark ($\downarrow$) & Canberra ($\downarrow$) & KL ($\downarrow$) & Cosine ($\uparrow$) & Intersection ($\uparrow$) & Chebyshev ($\downarrow$) & Clark ($\downarrow$) & Canberra ($\downarrow$) & KL ($\downarrow$) & Cosine ($\uparrow$) & Intersection ($\uparrow$) \\
            \midrule
            $\beta_{\text{low}}$ & 0.0368 & 0.2548 & 0.6477 & 0.0234 & 0.9782 & 0.9362 & 0.0794 & 0.6153 & 1.6458 & 0.1106 & 0.9167 & 0.8384 \\
            $\beta_{\text{mid}}$ & 0.0364 & 0.2519 & 0.6423 & 0.0227 & 0.9793 & 0.9374 & 0.0787 & 0.6121 & 1.6376 & 0.1092 & 0.9182 & 0.8403 \\
            $\beta_{\text{high}}$ & 0.0369 & 0.2576 & 0.6546 & 0.0243 & 0.9776 & 0.9358 & 0.0799 & 0.6184 & 1.6521 & 0.1114 & 0.9156 & 0.8372 \\
            $\{\beta_{\text{low}}, \beta_{\text{mid}}\}$ & 0.0353 & 0.2447 & 0.6245 & 0.0204 & 0.9824 & 0.9403 & 0.0774 & 0.6023 & 1.6102 & 0.1051 & 0.9221 & 0.8453 \\
            $\{\beta_{\text{mid}}, \beta_{\text{high}}\}$ & 0.0357 & 0.2468 & 0.6281 & 0.0209 & 0.9817 & 0.9392 & 0.0779 & 0.6051 & 1.6149 & 0.1062 & 0.9213 & 0.8441 \\
            $\{\beta_{\text{low}}, \beta_{\text{high}}\}$ & 0.0359 & 0.2492 & 0.6318 & 0.0216 & 0.9804 & 0.9383 & 0.0783 & 0.6082 & 1.6197 & 0.1073 & 0.9201 & 0.8426 \\
            $\mathcal{B} = \{\beta_{\text{low}}, \beta_{\text{mid}}, \beta_{\text{high}}\}$ & \textbf{0.0346} & \textbf{0.2369} & \textbf{0.6068} & \textbf{0.0177} & \textbf{0.9874} & \textbf{0.9427} & \textbf{0.0763} & \textbf{0.5932} & \textbf{1.5843} & \textbf{0.1009} & \textbf{0.9273} & \textbf{0.8496} \\
            \bottomrule
        \end{tabular}
    }
\end{table*}

\begin{figure}[!t]
\centering
\includegraphics[width=\linewidth]{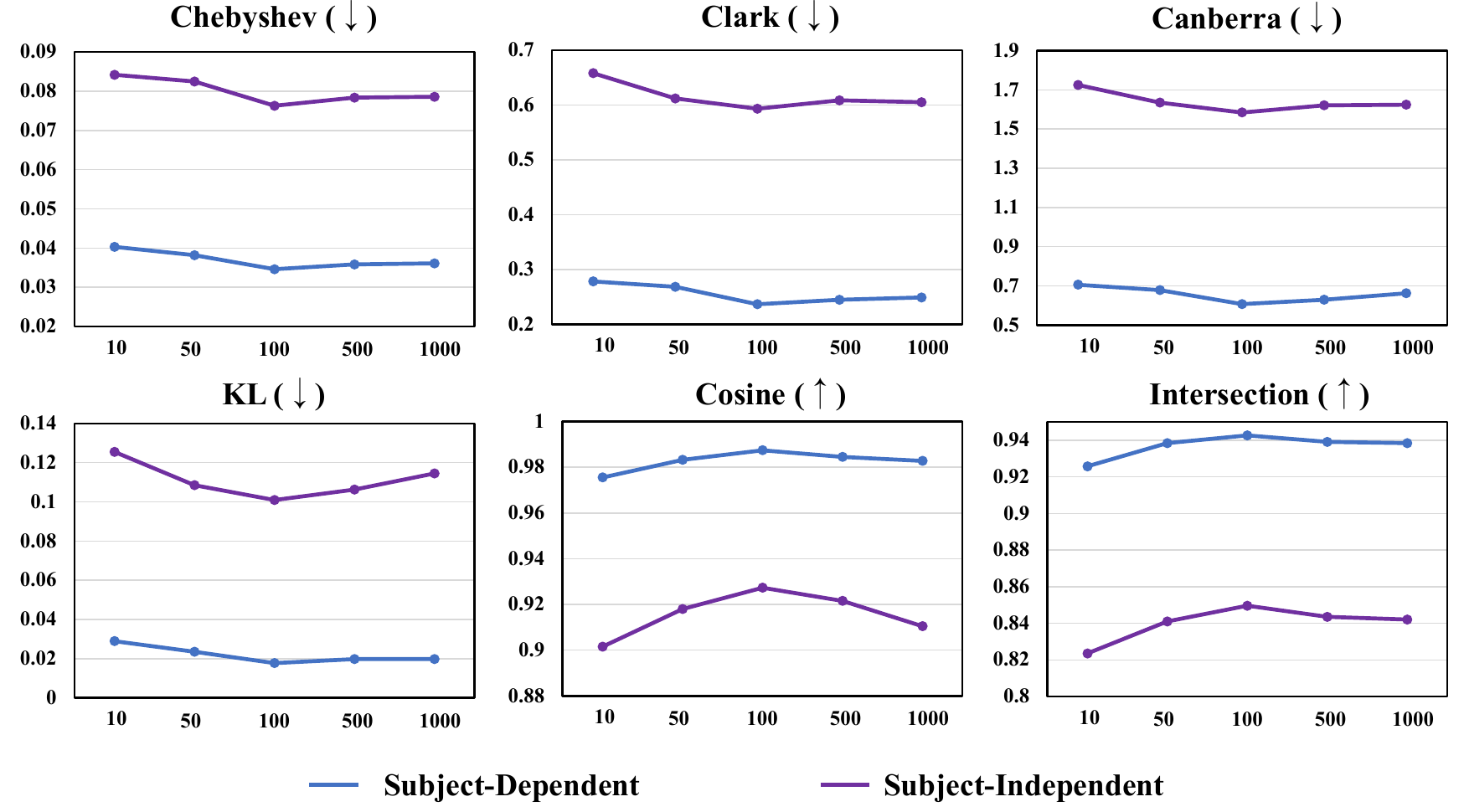}
\caption{Ablation study of the number of prototypes $M$ on the DMER dataset under subject-dependent and subject-independent settings.}
\vspace{-15pt}
\label{fig_number of M}
\end{figure}

\begin{figure}[!t]
\centering
\includegraphics[width=\linewidth]{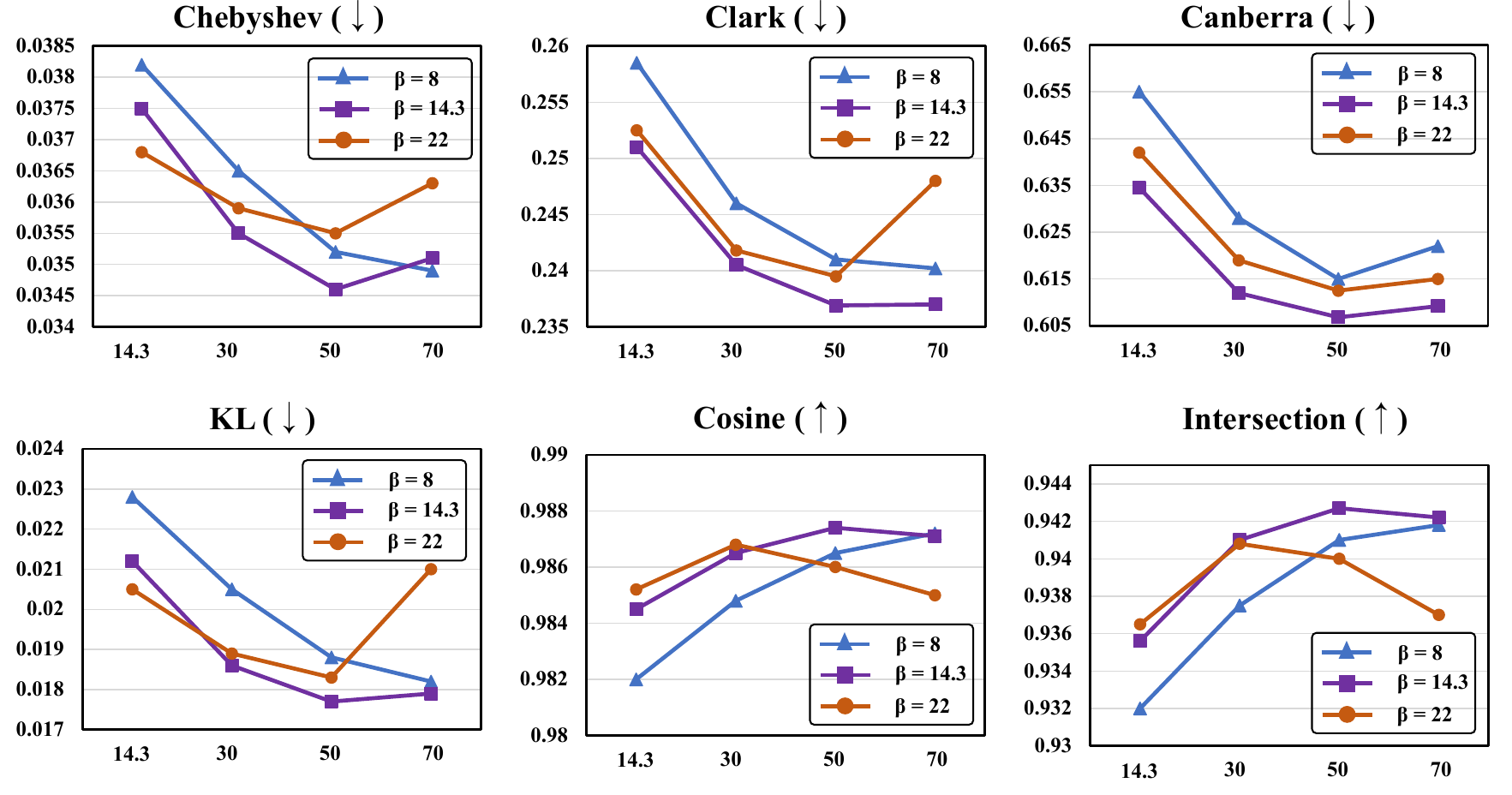}
\caption{Ablation study of the PCL module on the DMER dataset over different values of $\beta$ and $\tau_{\text{pcl}}^{-1}$. The horizontal axis shows $\tau_{\text{pcl}}^{-1}$; distinct curves correspond to different $\beta$ values.}
\label{fig_β}
\end{figure}

\begin{figure}[!t]
\centering
\includegraphics[width=\linewidth]{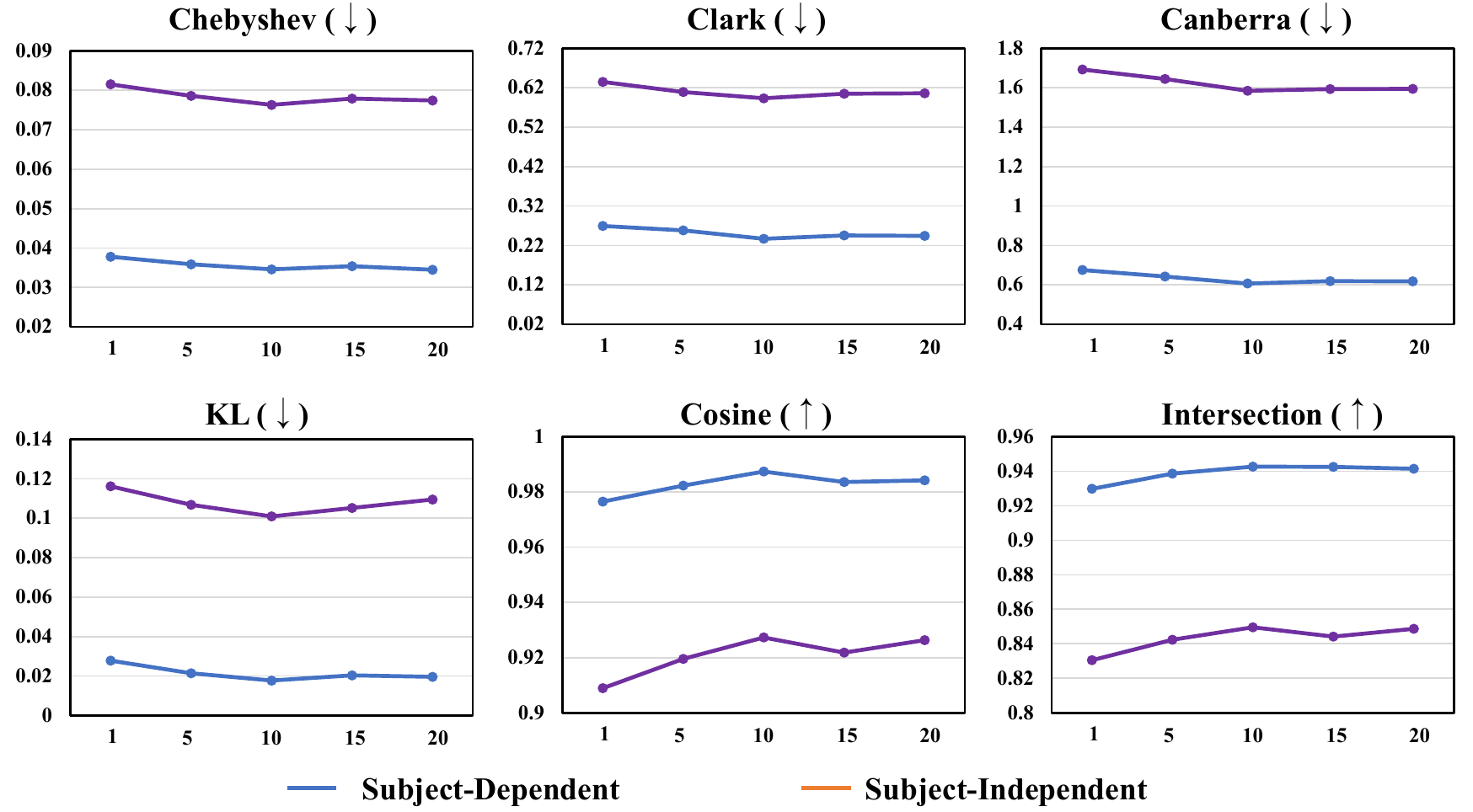}
\caption{Ablation study of the number of compression blocks $L$ in the HSC module on DMER under subject-dependent and subject-independent setting.}
\label{fig_number of L}
\end{figure}

We ablate key hyperparameters of MPCL on the DMER dataset.
\begin{itemize}
    \item Scales in MSAF: Table~\ref{tab:ablation_beta_dmer} shows that the three‑scale setting $\mathcal{B} = \{\beta_{\text{low}}, \beta_{\text{mid}}, \beta_{\text{high}}\}$ performs best. Smaller $\beta$ values promote global commonality aggregation, while larger ones emphasize local similarity.
    \item Number of prototypes $M$: Fig.~\ref{fig_number of M} indicates optimal performance at $M=100$. Too few prototypes cannot capture rich semantics, whereas too many lead to redundancy and sparsity.
    \item Parameters in PCL: Fig.~\ref{fig_β} shows the optimal combination $\beta=14.3$\footnote{The value 14.3 is the reciprocal of $\tau=0.07$, which is the commonly used parameter value in contrastive learning \cite{radford2021learning}.} and $\tau_{\text{pcl}}^{-1}=50$, among vary values $\beta\in(8, 14.3, 22)$ and $\tau_{\text{pcl}}^{-1}\in(14.3, 30, 50, 70)$.
    \item Number of compression blocks $L$: Fig.~\ref{fig_number of L} demonstrates that $L=10$ yields the best trade‑off between abstraction capacity and computational cost. Fewer layers under‑abstract features, while more layers offer diminishing returns.
\end{itemize}

\subsection{Visualization Analysis}

\begin{figure}[!t]
\centering
\includegraphics[width=\columnwidth]{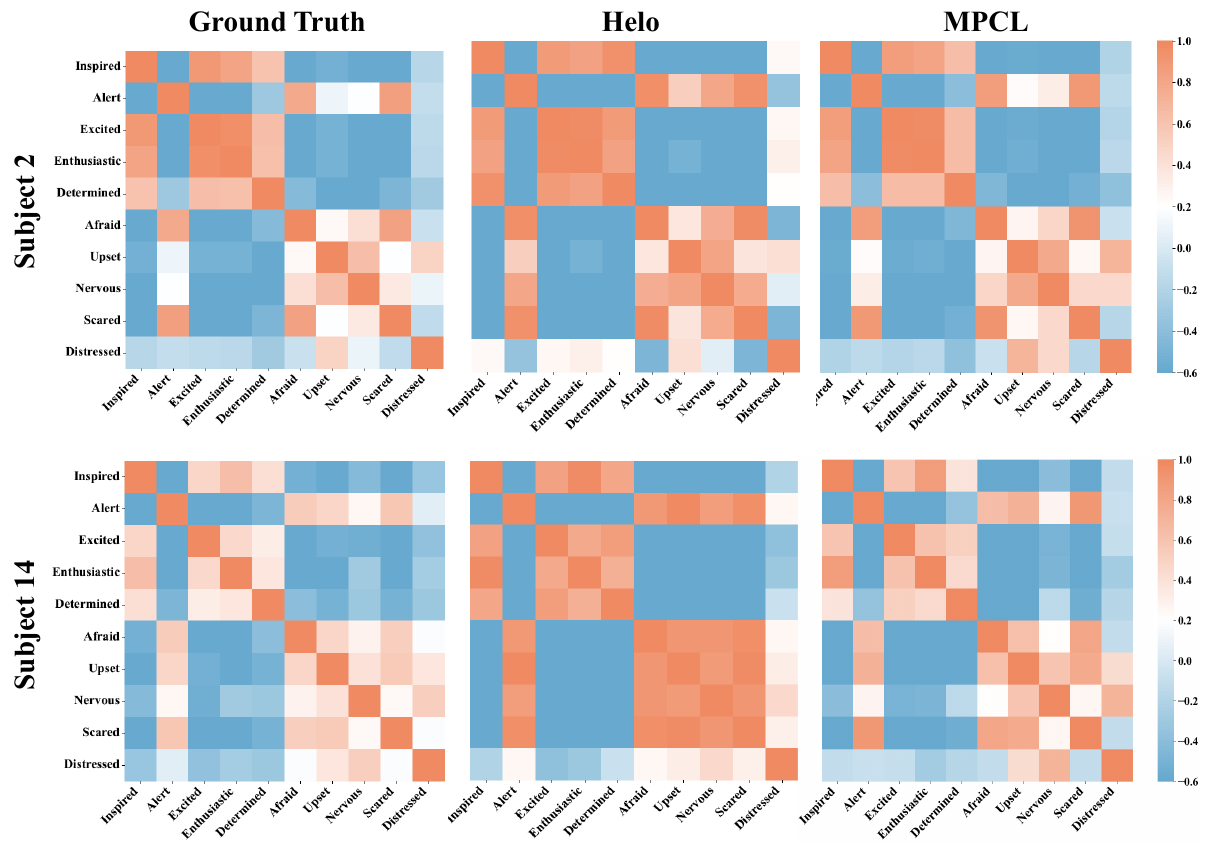}
\caption{Visualization of emotion‑label correlation matrices. Red intensity indicates the strength of correlation.}
\label{label}
\end{figure}

To examine the ability of MPCL in capturing complex co-occurrence relationships among emotions, we compute the Pearson correlation coefficients between predicted emotion distributions across all test samples and visualize the resulting label-correlation matrix. Fig.~\ref{label} presents a comparison for two subjects.

Compared to the HeLo method \cite{zheng2025helo}, the correlation matrix produced by MPCL aligns more closely with the ground truth (GT) in both pattern structure and intensity distribution. HeLo primarily models label dependencies by introducing learnable label embeddings and utilizing true label correlations as supervision signals. In contrast, MPCL leverages a Hopfield memory retrieval mechanism to amplify co-occurrence patterns among emotion prototypes, thereby learning semantic structural relations among emotions without explicit label‑correlation supervision. The results show that MPCL effectively recovers the intrinsic affective structure, highlighting its advantage in emotion distribution learning.

\section{Conclusion}

In this paper, we introduced the memory-guided prototypical co-occurrence learning (MPCL) framework, a novel approach that leverages associative memory mechanisms to mine prototype co‑occurrence patterns and semantic structural relationships in mixed‑emotion recognition. By employing associative memory networks, MPCL achieves multi-scale fusion of multimodal physiological data. To capture cross-modal semantic associations, we construct emotion prototype memory banks from both physiological and behavioral signals, producing semantically enriched representations through prototype‑weighted reconstruction. Inspired by the interplay between human perception and memory retrieval, we design a memory retrieval mechanism that reinforces frequently co‑occurring emotion prototypes, thereby modeling prototype co‑occurrence at a semantic level. Finally, a hierarchical semantic compression module further refines the affective representations, enabling accurate emotion‑distribution prediction. Extensive subject‑dependent and subject‑independent experiments on two public datasets (DMER and WESAD) demonstrate that MPCL consistently outperforms state‑of‑the‑art baselines in mixed‑emotion distribution learning.

\bibliographystyle{IEEEtran}
\bibliography{IEEEabrv,TPAMI}

\vfill

\end{document}